\documentclass[letterpaper, 10 pt, journal, twoside]{IEEEtran}
\IEEEoverridecommandlockouts                              

\usepackage{graphics} 
\usepackage{epsfig} 
\usepackage{amsmath} 
\usepackage{amssymb}  
\usepackage{tikz} 
\usetikzlibrary{shapes.geometric, arrows}
\usepackage{color} 
\usepackage{bm} 
\usepackage{gensymb}
\usepackage{dashbox}
\usepackage{marvosym}
\usepackage{cite} 
\usetikzlibrary{calc}
\usepackage{hyperref}
\usepackage[final]{changes} 
\newcommand{\expnumber}[2]{{#1}\mathrm{e}{#2}}

\title{Design and Control of a Variable Aerial Cable Towed System
}

\author{Zhen Li$^{1}$, Julian Erskine$^{2}$, St{\'e}phane Caro$^{3}$ and Abdelhamid Chriette$^{4}$
\thanks{$^{1}${\'E}cole Centrale de Nantes (ECN), Laboratoire des Sciences du Num{\'e}rique de Nantes (LS2N), UMR CNRS 6004, 1 rue de la Noe, 44321 Nantes, France
        {\tt\small zhen.li@eleves.ec-nantes.fr}}%
\thanks{$^{2}$ECN, LS2N
        {\tt\small julian.erskine@ls2n.fr}}%
\thanks{$^{3}$LS2N, Centre National de la Recherche Scientifique (CNRS), France
        {\tt\small stephane.caro@ls2n.fr}}%
\thanks{$^{4}$ECN, LS2N
        {\tt\small abdelhamid.chriette@ls2n.fr}}%
}

\begin{document}

\maketitle
 \thispagestyle{empty}
 \pagestyle{empty}

\begin{abstract}
Aerial Cable Towed Systems (ACTS) are composed of several Unmanned Aerial Vehicles (UAVs) connected to a payload by cables. Compared to towing objects from individual aerial vehicles, an ACTS has significant advantages such as heavier payload capacity, modularity, and full control of the payload pose. \added{They are however generally large with limited ability to meet geometric constraints while avoiding collisions between UAVs.} This paper presents \added{the modelling, performance analysis, design, and a proposed controller for} a novel ACTS with variable cable lengths, named Variable Aerial Cable Towed System (VACTS).

Winches are embedded on the UAVs for actuating the cable lengths similar to a Cable-Driven Parallel Robot to increase the versatility of the ACTS. The general geometric, kinematic and dynamic models of the VACTS are derived, followed by the development of a centralized feedback linearization controller. The  design is based on a wrench analysis of the VACTS, without \added{constraining the cables to pass through the UAV center of mass, as in current works}. Additionally, the performance of the VACTS and ACTS are compared \added{showing that the added versatility comes at the cost of payload and configuration flexibility}. A prototype confirms the feasibility of the system.
\end{abstract}


\section{INTRODUCTION}

\IEEEPARstart{C}{ombining} the agility of aerial vehicles with the manipulation capability of manipulators makes aerial manipulation an attractive topic\cite{literature}. Unmanned Aerial Vehicles (UAVs) are becoming more and more popular during last decades, not only in research fields but also in commercial applications. Amongst UAVs, quadrotors are widely used in photographing, inspection, transportation and (recently) manipulation because of their agility, versatility, and low cost. However, the coupling between their translational and rotational dynamics complicates tasks requiring fine pose control. Moreover, most have limited payload capacity, which makes it difficult to transport large objects by an individual quadrotor. Multi-quadrotor collaboration is studied as a solution to compensate the previous two drawbacks inherent to individual quadrotors. 

In this paper, we study a payload suspended via cables, which reduce the couplings between the platform and quadrotors attitude dynamics, allow reconfigurability, and reduce aerodynamic interference between the quadrotor downwash and the payload. Moreover, Cable-Driven Parallel Robots (CDPRs) are well-studied,  providing a part of theoretical support, such as tension distribution algorithms in \cite{bookCDPR}, \cite{li2013optimal}, \cite{lamaury2013tension},~\cite{gouttefarde2015versatile}, controllers \cite{RCDPR}, \cite{adaptive1} and wrench performance evaluations \cite{wrenchbasic}. Reconfigurable CDPRs with both continuously moving \cite{RCDPR, fastkit} and discrete \cite{DRCDPR} cable anchor points are proposed to solve problems such as wrench infeasiblility and self or environmental collisions.

For the past ten years, the Aerial Cable Towed System (ACTS) composed of several UAVs, a payload, and cables to cooperatively manipulate objects, has attracted researchers' attention and has been well developed from a controls viewpoint. An ACTS prototype for 6-dimensional manipulation "FlyCrane" and a motion planning approach called Transition-based Rapidly-exploring Random Tree (T-RRT) are proposed in \cite{flycrane}. In \cite{cone} an ACTS prototype for 3-dimensional manipulation and a decentralized linear quadratic control law are proposed and \cite{RCDPR} develops a general feedback linearization control scheme for over-actuated ACTS with a 6-DOF payload. An ACTS prototype with three quadrotors and a point mass is implemented in \cite{lastone} and the performance was evaluated using capacity margin, a wrench-based robustness index adapted from the CDPR in \cite{wrench}.

\begin{figure}[!tpb]
      \centering
      \includegraphics[width=1\linewidth]{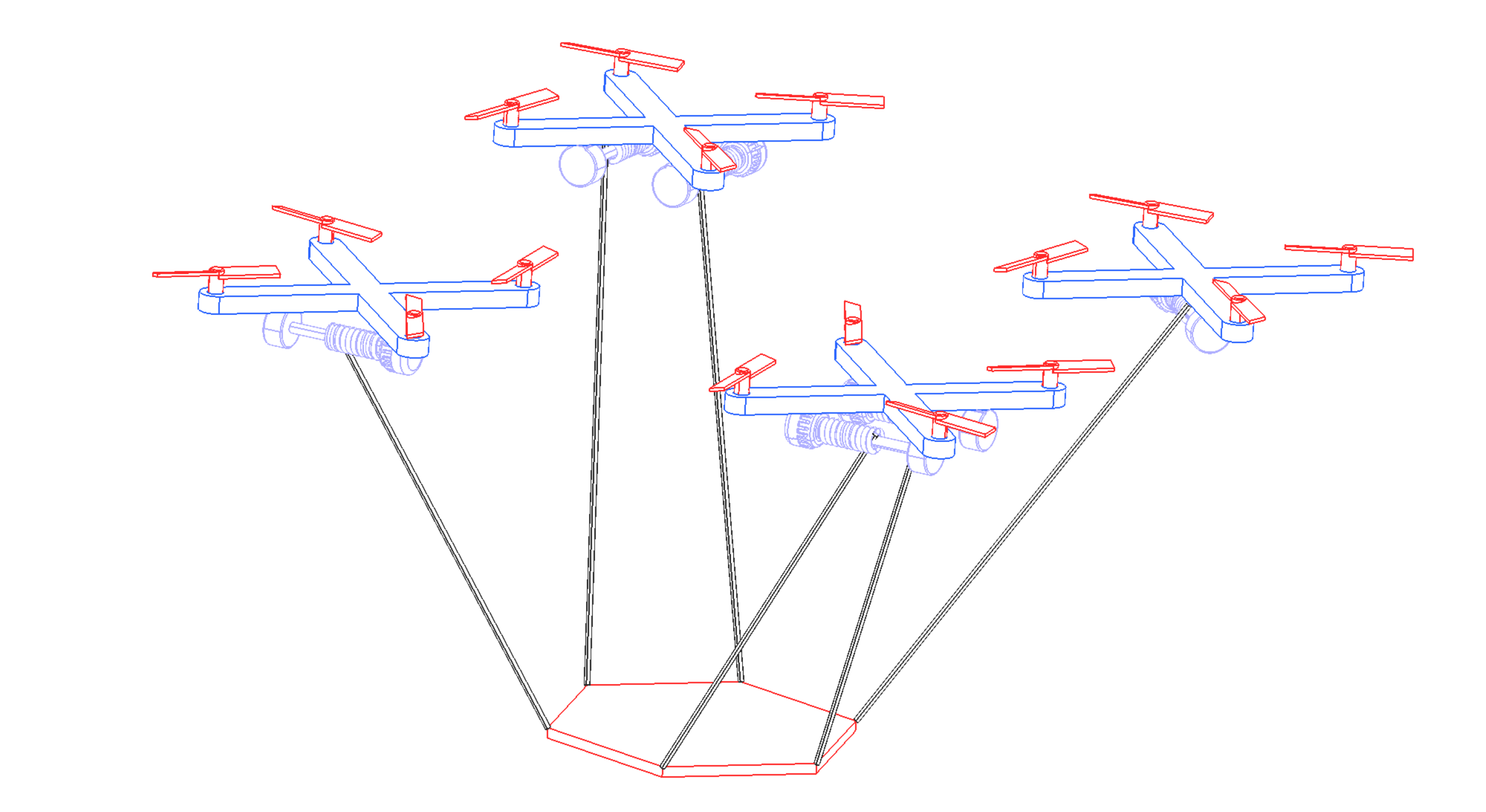}
      \caption{The sketch structure of a VACTS with 4 quadrotors, 6 winches and cables, and a platform}
      \label{fig:acts}
\end{figure}

The ACTS shows promise both in aerial manipulation and sharing heavy burden, however there are still some limitations for current designs.\added{ The size of ACTS is usually large. As a consequence, it is not suitable for motion in cluttered environments. Besides, short cables may lead to self collisions, particularly on takeoff where cable tension discontinuities may lead to poor control. Furthermore, the ACTS may have to change the distribution of UAVs for motion in a cluttered environment, which may lead to wrench infeasibility and self collisions.} Therefore, a novel Aerial Cable Towed System with actuated cable lengths, the Variable Aerial Cable Towed System (VACTS) is proposed in this paper to make up for these shortcomings. The actuated cable lengths can reshape the size of overall system, which implies the possibility of passing through a constrained environment or limited space\footnote{\href{https://drive.google.com/file/d/1IO3qvFyWSTnUMLefXeysFpoqPMR585ud/view?usp=sharing}{\color{blue}{\nolinkurl{https://drive.google.com/file/d/1IO3qvFyWSTnUMLefXeysFpoqPMR585ud/view?usp=sharing}}}}. \added{It also allows flexibility in choosing between external load resistance and energy efficiency for some systems, with the possibility of modifying the wrench capabilities of certain ACTS systems such as the Flycrane.} Moreover, it is conceivable that combining force control of the quadrotors with velocity control by actuating cable lengths might improve payload positioning precision. 

\added{This  modification to the common ACTS is first proposed in this paper, and a general model is rigorously built with few assumptions. A prototype of this novel system (to our knowledge, the first ACTS with actuated cables) is presented in this paper, showing the system is feasible.} Section~\ref{sec:model} derives the geometric, kinematic and dynamic models. Section~\ref{sec:control} illustrates the architecture of a centralized feedback linearization controller. The design of the VACTS and its performance relative to the ACTS are compared in Sec.~\ref{sec:wrench}. Section~\ref{sec:exp} discusses the experimental results. Conclusions and future work are presented in Sec.~\ref{sec:con}. 

\section{Modelling}\label{sec:model}

The geometric, kinematic and dynamic models of the VACTS are derived in this section. The sketch structure of a VACTS is shown in Fig.~\ref{fig:acts}. It is composed of $n$ quadrotors, $m$  cables  and winches ($m\geq n$), and a payload. There are $s_j \in [1,2]$ winches mounted on the $j^{\text{th}}$ quadrotor. \added{Unlike existing ACTS models where the cable passes through the quadrotor center of mass (COM), each winch of the VACTS may impose geometric constraints, therefor the model  considers the cable attached to an arbitrary point on the quadrotor.}

\subsection{Geometric Modelling of the VACTS}

\begin{figure}[!tpb]
   	\centering
   	\begin{tikzpicture}
   	\node[anchor=south west,inner sep=0] (image) at(0,0){\includegraphics[width=\linewidth]{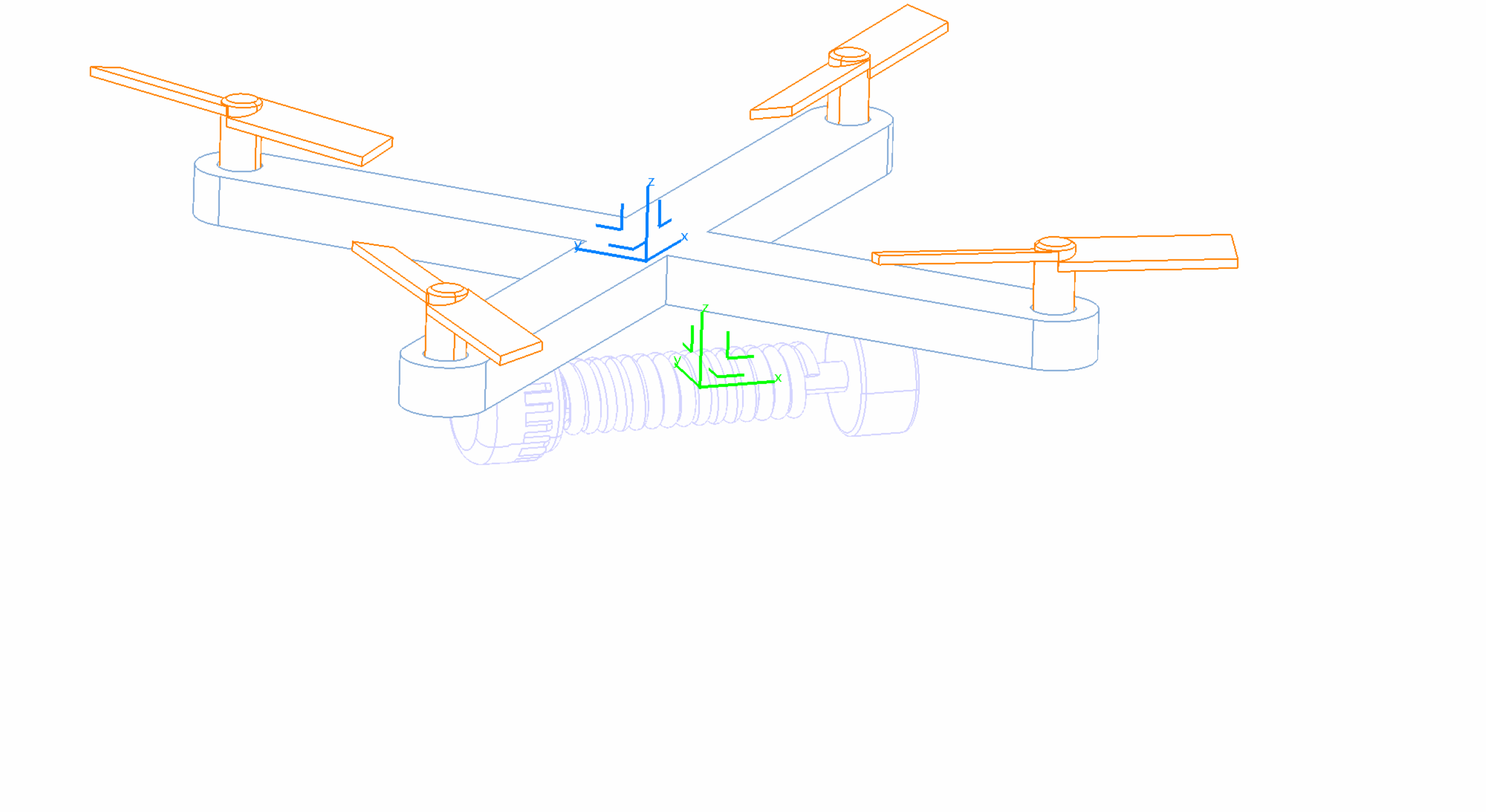}};
   \begin{scope}[x={(image.south east)},y={(image.north west)}]
   
   \draw[fill=gray,fill opacity=0.1,gray](.65, .15)--(0.95,0.45)..controls (0.9,0.35) and (0.99,0.2)..(0.97,0.1)--(.65, .15);
   \draw[fill=gray,fill opacity=0.1,gray](.65, .15)--(0.97,0.1)--(0.97,0.05)--(.65, .1)--(.65, .15);
   
   \draw[->,>=angle 60,line width=.5pt](0.1, 0.1)--node [above left] {$x_0$}(.05, .05);
   \draw[->,>=angle 60,line width=.5pt](0.1, 0.1)--node [above right] {$y_0$}(.2, .1);
   \draw[->,>=angle 60,line width=.5pt](0.1, 0.1)--node [above right] {$z_0$}(.1, .2);
   \draw (.15, .4) node [below] {$\mathcal{F}_0$};
   \draw[fill] (.1,.1)circle(0.05cm) node [below right] {$O$};
   
   \draw[blue] (.44, .8) node [above] {$\mathcal{F}_j$};
   \draw[blue,fill] (.435, .68)circle(0.05cm) node [above left] {$J$};
   
   \definecolor{mygreen}{rgb}{0,0.6,0.3};
   \draw[mygreen] (.5, .65) node {$\mathcal{F}_{w_i}$};
   \draw[mygreen,fill] (.47, .52) circle(0.05cm) node [below left] {$W_i$};
   \draw[mygreen,fill] (.552, .56) circle(0.05cm) node [right] {$I_i$};
   
   \draw[red,->,>=angle 60,line width=.5pt](.9, .3)--node[xshift=-0.3cm, yshift=-0.1cm]{$x_p$}(.85, .25);
   \draw[red,->,>=angle 60,line width=.5pt](.9, .3)--node[above right]{$y_p$}(.97, .3);
   \draw[red,->,>=angle 60,line width=.5pt](.9, .3)--node[above]{$z_p$}(.9, .4);
   \draw[red] (.9, .6) node [below] {$\mathcal{F}_p$};
   \draw[red,fill] (.9, .3) circle(0.05cm) node [below right] {$P$};
   \draw[->,>=angle 60,line width=1pt](.05,.5)--node [above left] {$\bm{g}$} (.05,.3);
   \draw[red,line width=.5pt](.65, .15)--node [above right] {$l_i$}(.552, .56);
   \draw[red,fill] (.65, .15) circle(0.05cm) node [below] {$B_i$};
   
   \draw[dashed,->,>=angle 60](.65, .15)--node[above right]{$z_p$}(.65, .35);
   \draw[dashed,->,>=angle 60](.65, .15)--node[yshift=-0.2cm]{$x_p$}(.55, .1);
   \draw[dashed](.552, .56)--(.552, .25);
   \draw[dashed](.552, .25)--(.65, .15);
   \draw(.552,.27)--(.56,.26);
   \draw(.56,.26)--(.56,.24);
   \draw[red,->,>=angle 60](.65, .25) to [out=180,in=60] node [xshift=-0.05cm, yshift=0.25cm]{${\theta}_i$} (.63, .22);
   \draw[red,->,>=angle 60](.6, .12) to [out=110,in=-130] node [left]{${\phi}_i$} (.61, .2);
   \draw[red,->,>=angle 60,line width=1pt](.6,.3)--node [left] {${}^{p}{\bm{u}}_i$} (.58,.38);
   
   \draw[blue,->,>=angle 60,line width=1pt](.1,.1) to [out=110,in=170] node [left]{$^0{\bm{T}}_j$} (.435, .68);
   \draw[mygreen,->,>=angle 60,line width=.5pt](.435, .68) to [out=-90,in=140] node [left]{$^j{\bm{T}}_{w_i}$} (.47,.52);
   \draw[red,->,>=angle 45,line width=1pt](.1,.1) to [out=-10,in=-135] node [xshift=-1.5cm,yshift=0.4cm]{$^0{\bm{T}}_p$} (.9,.3);
   \draw[mygreen,->,>=angle 45,line width=.5pt](.47,.52) to [out=0,in=-140] node [below]{$^w{\bm{x}}_{Ii}$} (.552, .56);
   \draw[red,->,>=angle 60,line width=1pt](.9,.3) to [out=180,in=30] node [above]{$^p{\bm{x}}_{B_i}$} (.65, .15);
   \draw (.57,.65) circle (0.1cm) node [above right] {$G$};
   \fill (.57,.65) -- (.585,.65) arc (0:90:0.1cm);
   \fill (.57,.65) -- (.555,.65) arc (180:270:0.1cm);
   \draw (.8,.2) circle (0.1cm) node [below left] {$C$};
   \fill (.8,.2) -- (.815,.2) arc (0:90:0.1cm);
   \fill (.8,.2) -- (.785,.2) arc (180:270:0.1cm);
\normalsize    
 \end{scope}  
 \end{tikzpicture}
    \vspace{-3em}\caption{\added{Parametrization of  VACTS with the quadrotor, winch and platform}}
  	\label{fig:actsparam}
   	\end{figure}

The geometric parametrization is presented in Fig.~\ref{fig:actsparam} with symbolic interpretation listed in Table~\ref{table:param}. Matrix  $^a{\bm{T}}_b$ is the homogeneous transformation matrix from frame $\mathcal{F}_a$ to $\mathcal{F}_b$, and consists of a rotation matrix ${}^a{\bm{R}}_b$ and a translation vector ${}^a{\bm{x}}_b$. Matrix $^j{\bm{T}}_{wi}$ expresses the transform of the $i^{\text{th}}$ winch from the $j^{\text{th}}$ quadrotors frame $\mathcal{F}_j$ located at it's center of mass (COM). The cable length is $l_i$, and the unit vector along cable direction is expressed in the payload frame $\mathcal{F}_p$ as ${}^p{\bm{u}}_i={\begin{bmatrix}
 c_{{\phi}_i}s_{{\theta}_i} &
 s_{{\phi}_i}s_{{\theta}_i} &
 c_{{\theta}_i}
 \end{bmatrix}}^T$ via azimuth angle ${\phi}_i$ and inclination angle ${\theta}_i$, as used in \cite{lastone}. The gravity vector is $\bm{g}=[0 \; 0 \; -9.81]\text{~ms}^{-2}$. \added{Note that if $s_j = 2$, the two cables share a coupled motion wrt the payload.}

The $i^{\text{th}}$ loop closure equation can be derived considering the forward and backward derivation of ${}^0{\bm{x}}_{Ii}$.
\begin{equation}\label{eq:GM}
{}^0{\bm{x}}_p + {}^0{\bm{R}}_p {}^p{\bm{x}}_{B_i} + l_i {}^0{\bm{R}}_p {}^p{\bm{u}}_i = {}^0{\bm{x}}_j + {}^0{\bm{R}}_j {}^j{\bm{x}}_{Ii}
\end{equation}

\subsection{Kinematic Modelling of the VACTS} 

After differentiating (\ref{eq:GM}) with respect to time, the first order kinematic model for each limb can be derived as~(\ref{eq:KM}), where the payload has translational velocity ${}^0{\dot{\bm{x}}}_p$ and angular velocity ${}^0{\bm{\omega}}_p$, and the $j^{\text{th}}$ quadrotor has translational velocity ${}^0{\dot{\bm{x}}}_j$ and angular velocity ${}^0{\bm{\omega}}_j$.
\begin{equation}\label{eq:KM}\footnotesize
\begin{aligned}
{}^0{\dot{\bm{x}}}_{p} + {}^0{\bm{\omega}}_p \times ({}^0{\bm{R}}_p {}^p{\bm{x}}_{B_i}) + l_i {}^0{\bm{R}}_p {}^p{\dot{\bm{u}}}_i + l_i {}^0{\bm{\omega}}_p \times ({}^0{\bm{R}}_p {}^p{{\bm{u}}}_i) \\
= {}^0{\dot{\bm{x}}}_{j} - {\dot{l}}_i {}^0{\bm{R}}_p {}^p{\bm{u}}_i + {}^0{\bm{\omega}}_j \times ({}^0{\bm{R}}_j {}^j{\bm{x}}_{Ii}) + {}^0{\bm{R}}_j {}^j{\dot{\bm{x}}}_{Ii}
\end{aligned}\normalsize
\end{equation}

\begin{table}[!tpb]
\caption{\added{Nomenclature}}
\label{table:param}
\begin{center}
\vspace{-1em}\begin{tabular}{|c||c|}
  \hline
  symbol & physical meaning \\
  \hline
  $O$ & Origin of world frame $\mathcal{F}_0$ \\
  $J$ & The $j^{\text{th}}$ quadrotor centroid and origin of $\mathcal{F}_j$ \\
  $W_i$ & The $i^{\text{th}}$ winch centroid \\
  $I_i$ & Coincident point of the $i^{\text{th}}$ cable-winch pair \\
  $B_i$ & Attachment point of the $i^{\text{th}}$ cable-payload pair \\
  $P$ & Origin of payload frame $\mathcal{F}_p$ \\
  $C$ & COM of the payload \\
  $G$ & COM of the $j^{\text{th}}$ quadrotor with embedded winches \\
  \hline
\end{tabular}
\end{center}
\end{table}

The first order kinematic model can be re-expressed as~(\ref{eq:1KM}), where the Jacobian matrix ${\bm{J}}_{(6+3m) \times 3n}$ can be obtained from the forward Jacobian matrix ${\bm{A}}_{3m \times (6+3m)}$ and the inverse Jacobian matrix ${\bm{B}}_{3m \times 3n}$. ${\bm{A}}^{+}$ is the pseudo-inverse of $\bm{A}$, which minimizes the 2-norm of  ${\dot{\bm{x}}}_{\bm{t}}$ since the system is under-determined. For example, if $s_1=1$ and $s_2=2$, the matrix $\bm{B}$ will be expressed as (\ref{eq:B}). The task space vector is ${\bm{x_t}}_{(6+3m) \times 1}$ and the joint space vector is ${\bm{q_a}}_{3n \times 1}$. Note that ${\dot{\bm{x}}}_{\bm{t}}$ is not the  time derivative of $\bm{x_t}$ considering that the angular velocity ${}^0{\bm{\omega}}_p$ is not the time derivative of payload orientation, which can be expressed in the form of Euler angles or quaternions. The form ${[\bm{x}]}_{\times}$ is the cross product matrix of vector ${\bm{x}}$ and ${\bf{I}}_3$ is a $3\times 3$ identity matrix. 
\begin{equation}\label{eq:1KM}
{\dot{\bm{x}}}_{\bm{t}} = \bm{J} {\dot{\bm{q}}}_{\bm{a}} + \bm{a} \text{, with } \bm{J} = {\bm{A}}^{+} \bm{B}
\end{equation}
\begin{equation}\label{eq:Xdot}\small
{\dot{\bm{x}}}_{\bm{t}}={\begin{bmatrix} {}^0{\dot{\bm{x}}}_{p}^T & {}^0{\bm{\omega}}_p^T & {\dot{\phi}}_1 & {\dot{\theta}}_1 & {\dot{l}}_1 & \cdots & {\dot{\phi}}_m & {\dot{\theta}}_m & {\dot{l}}_m \end{bmatrix}}^T \normalsize
\end{equation}
\begin{equation}\label{eq:qadot}
{\dot{\bm{q}}}_{\bm{a}}={\begin{bmatrix} {}^0{\dot{\bm{x}}}_{1}^T & \cdots & {}^0{\dot{\bm{x}}}_{n}^T \end{bmatrix}}^T
\end{equation}
\begin{equation}\label{eq:A}\footnotesize \setlength{\arraycolsep}{1pt}
\begin{aligned}
\bm{A}=&\begin{bmatrix} 
{\mathbb{I}}_3 & {\bm{A}}_{12} & l_1 {}^0{\bm{R}}_p {\bm{C}}_1 & {}^0{\bm{R}}_p {}^p{\bm{u}}_1 \\
\vdots & \vdots &&& \ddots& \ddots\\
{\mathbb{I}}_3 & {\bm{A}}_{m2} &&&&& l_m {}^0{\bm{R}}_p {\bm{C}}_m & {}^0{\bm{R}}_p {}^p{\bm{u}}_m
\end{bmatrix} \\
&\text{with } {\bm{A}}_{i2} = {[{}^0{\bm{R}}_p ({}^p{\bm{x}}_{B_i} + l_i {}^p{\bm{u}}_i)]}_{\times}^T
\end{aligned}\normalsize 
\end{equation}
\begin{equation}\label{eq:B}
\bm{B}=\begin{bmatrix}
{\bf{I}}_3 \\
& {\bf{I}}_3 \\
& {\bf{I}}_3 \\
&& \ddots\\
&&& {\bf{I}}_3
\end{bmatrix} \begin{matrix}
\Rightarrow s_1 \\ \\
\Rightarrow s_2 \\ 
\vdots \\
\Rightarrow s_n
\end{matrix}
\end{equation}
\begin{equation}\label{eq:C}
{\bm{C}}_i = \begin{bmatrix}
-s_{{\phi}_i}s_{{\theta}_i} & c_{{\phi}_i}c_{{\theta}_i} \\
c_{{\phi}_i}s_{{\theta}_i} & s_{{\phi}_i}c_{{\theta}_i} \\
0 & -s_{{\theta}_i}
\end{bmatrix}
\end{equation}
\begin{equation}\label{eq:a}\\
\setlength{\arraycolsep}{1pt}
\bm{a} = {\bm{A}}^{+} \begin{bmatrix}
{}^0{\bm{\omega}}_1 \times ({}^0{\bm{R}}_1 {}^1{\bm{x}}_{I_1}) + {}^0{\bm{R}}_1 {}^1{\dot{\bm{x}}}_{I_1} \\
{}^0{\bm{\omega}}_2 \times ({}^0{\bm{R}}_2 {}^2{\bm{x}}_{I_2}) + {}^0{\bm{R}}_2 {}^2{\dot{\bm{x}}}_{I_2} \\
\vdots \\
{}^0{\bm{\omega}}_n \times ({}^0{\bm{R}}_n {}^n{\bm{x}}_{I_m}) + {}^0{\bm{R}}_n {}^n{\dot{\bm{x}}}_{I_m}
\end{bmatrix}
\end{equation}

Additionally, the second order kinematic model is also derived in matrix form as (\ref{eq:2KM}), where $\bm{b}$ is the component related to the derivatives of $\bm{A}$ and $\bm{B}$. ${\bm{B}}^{+}$ is the pseudo-inverse of $\bm{B}$, which will minimize the residue of the equation system if the system is over-determined, i.e. $m>n$.
\begin{equation}\label{eq:2KM}
{\ddot{\bm{q}}}_{\bm{a}} = {\bm{B}}^{+} \bm{A} {\ddot{\bm{x}}}_{\bm{t}} + {\bm{B}}^{+} \bm{b} \text{, with }\bm{b}={\begin{bmatrix} {\bm{b}}_1^T & \cdots & {\bm{b}}_m^T \end{bmatrix}}^T
\end{equation}
\begin{equation}\label{eq:bi}
\small \setlength{\arraycolsep}{2pt}
\begin{aligned}
{\bm{b}}_i= & {}^0{\bm{\omega}}_p \times ({}^0{\bm{\omega}}_p \times {}^0{\bm{R}}_p ({}^p{\bm{x}}_{B_i}+l_i {}^p{{\bm{u}}}_i)) \\
& + 2{\dot{l}}_i {}^0{\bm{\omega}}_p \times ({}^0{\bm{R}}_p {}^p{{\bm{u}}}_i) + 2{\dot{l}}_i {}^0{\bm{R}}_p {\bm{C}}_i {\begin{bmatrix}
{\dot{\phi}}_i & {\dot{\theta}}_i
\end{bmatrix}}^T \\
& + 2 l_i {}^0{\bm{\omega}}_p \times ({}^0{\bm{R}}_p {\bm{C}}_i {\begin{bmatrix}
{\dot{\phi}}_i & {\dot{\theta}}_i \end{bmatrix}}^T) + l_i {}^0{\bm{R}}_p {\dot{\bm{C}}}_i {\begin{bmatrix}
{\dot{\phi}}_i & {\dot{\theta}}_i
\end{bmatrix}}^T \\
& - {}^0{\dot{\bm{\omega}}}_j \times ({}^0{\bm{R}}_j {}^j{\bm{x}}_{Ii}) - {}^0{\bm{\omega}}_j \times ({}^0{\bm{\omega}}_j \times {}^0{\bm{R}}_j {}^j{\bm{x}}_{Ii}) \\
& -2{}^0{\bm{\omega}}_j \times {}^0{\bm{R}}_j {}^j{\dot{\bm{x}}}_{Ii} - {}^0{\bm{R}}_j {}^j{\ddot{\bm{x}}}_{Ii}
\end{aligned}\normalsize
\end{equation}

\subsection{Dynamic Modelling of the VACTS}

\subsubsection{For the payload}

By using the Newton-Euler formalism, the following dynamic equation is presented, taking into account of an external wrench ${\bm{w}}_e$ \added{exerted on the payload by the  environment}, consisting of force ${\bm{f}}_e$ and moment ${\bm{m}}_e$. \added{Along with ${\bm{w}}_e$, inertial forces from the mass $m_p$ and inertia matrix ${\bm{I}}_p$ of the payload, and cable tensions $t_i {}^p{\bm{u}}_i$  act on the payload}. Eqns.~(\ref{eqn:platform_translation_dynamics},\ref{eqn:platform_rotation_dynamics}) express the dynamics of the platform.
\begin{equation}\label{eqn:platform_translation_dynamics}
 {\bm{f}}_e + m_p\bm{g}+{}^0{\bm{R}}_p \sum_{i=1}^{m} t_i {}^p{\bm{u}}_i = m_p {}^0{\ddot{\bm{x}}}_p
 \end{equation}
 \begin{equation}\label{eqn:platform_rotation_dynamics}
 \begin{aligned}
 {\bm{m}}_e + ({}^0{\bm{R}}_p {}^p{\bm{x}}_C) \times m_p\bm{g} + {}^0{\bm{R}}_p \sum_{i=1}^{m}{}^p{\bm{x}}_{B_i} \times t_i {}^p{\bm{u}}_i \\
 ={\bm{I}}_p {}^0{\dot{\bm{\omega}}}_p + {}^0{\bm{\omega}}_p \times ({\bm{I}}_p {}^0{\bm{\omega}}_p)
 \end{aligned}
 \end{equation}
 
 If these equations are expressed in matrix form, the cable tension vector $\bm{t}={\begin{bmatrix}t_1 & \cdots & t_m\end{bmatrix}}^T$ can be derived as (\ref{eq:DM1}) considering a wrench matrix $\bm{W}$ and a mass matrix ${\bm{M}}_p$. ${\bm{W}}^{+}$ is the pseudo-inverse of $\bm{W}$ and $\mathcal{N}(\bm{W})$ represents the null-space of $\bm{W}$, used in some tension distribution algorithms \added{if the payload is over-actuated} \cite{lamaury2013tension},~\cite{gouttefarde2015versatile}.
 \begin{equation}\label{eq:DM1}
 \bm{t} = {\bm{W}}^{+} \left( {\bm{M}}_p {\begin{bmatrix}
 {}^0{\ddot{\bm{x}}}_p^T & {}^0{\dot{\bm{\omega}}}_p^T
\end{bmatrix}}^T + \bm{c} \right) + \mathcal{N}(\bm{W})
 \end{equation}
 \begin{equation}
 {\bm{M}}_p = \begin{bmatrix}
 m_p {\mathbb{I}}_3 & {\bm{0}} \\
 {\bm{0}} & {\bm{I}}_p
 \end{bmatrix}    
 \end{equation}
 \begin{equation}\label{eq:W}
 \setlength{\arraycolsep}{2pt}
 \bm{W}=\begin{bmatrix}
 {{}^0{\bm{R}}_p {}^p\bm{u}}_1 &\cdots& {}^0{\bm{R}}_p {}^p{\bm{u}}_m\\
 {}^0{\bm{R}}_p ({}^p{\bm{x}}_{B_1}\times {}^p{\bm{u}}_1) &\cdots& {}^0{\bm{R}}_p ({}^p{\bm{x}}_{B_m}\times {}^p{\bm{u}}_m)
 \end{bmatrix}
 \end{equation}
 \begin{equation}\label{eq:c}
\setlength{\arraycolsep}{2pt}
  \bm{c}= \begin{bmatrix}
 {\bm{0}}\\ 
 {}^0{\bm{\omega}}_p \times ({\bm{I}}_p {}^0{\bm{\omega}}_p)
\end{bmatrix} - \begin{bmatrix}
m_p\bm{g} \\ ({}^0{\bm{R}}_p {}^p{\bm{x}}_C) \times m_p\bm{g}
\end{bmatrix} - {\bm{w}}_e
 \end{equation}
 
\subsubsection{For the winch}

All winches are assumed to have the same design, specifically the drum radius of all winches is $r_{d}$, the mass is $m_w$, and the inertia matrix is ${\bm{I}}_w$. The $x$-axis of the winch frame is along the winch drum \added{axis of rotation}. The rotational rate of the $i^{\text{th}}$ winch is ${\omega}_{r_i}$. It relates to the change rate of cable length as (\ref{eq:lengthrate}) considering that the radius of cable is small \added{(reasonable, as payloads are generally light)}.
\begin{equation}\label{eq:lengthrate}
  {\dot{l}}_i= r_d {\omega}_{r_i}
  \end{equation}
  
By using the Newton-Euler formalism, the following dynamic equation is expressed in $\mathcal{F}_{wi}$, where the \added{winch} torque is ${\tau}_i$. Additionally, we define a constant vector $\bm{k}=\begin{bmatrix} 1&0&0 \end{bmatrix}$ to represent the torque generated from cable tension orthogonal to the drum surface. 
\begin{equation}
 {{\tau}}_{i} - \bm{k} ({}^w{\bm{x}}_{Ii} \times t_i{}^w{\bm{u}}_i) = {{I}}_{xx} {\dot{\omega}}_{r_i}
 \end{equation}

\subsubsection{For the quadrotor}
 
By using the Newton-Euler formalism, the following dynamic equations can be derived taking into account of the thrust force ${\bm{f}}_j$ and thrust moment ${\bm{m}}_j$ generated by the quadrotor's motors. The mass of the $j^{\text{th}}$ quadrotor with $s_j$ embedded winches is $m_j= m_q + s_j m_w$ and the inertia matrix  ${\bm{I}}_j$\added{, is from the parallel axis theorem}.
\begin{equation}\label{eq:quadforce}
  {\bm{f}}_j + m_j\bm{g} - {}^0{\bm{R}}_p \sum_{k=1}^{s_j} t_{k} {}^p{\bm{u}}_{k} = m_{j} {}^0{\ddot{\bm{x}}}_j
 \end{equation}
 \begin{equation}\label{eq:quadmoment}
 \begin{aligned}
 {\bm{m}}_j + \underbrace{({}^0{\bm{R}}_j{}^j{\bm{x}}_G) \times m_j \bm{g}}_{\text{gravity}} - {}^0{\bm{R}}_j \sum_{k=1}^{s_j} \underbrace{{}^j{\bm{x}}_{I_{k}} \times t_{k} {}^p{\bm{u}}_{k}}_{\text{cable  tension}} \\
  =  {\bm{I}}_j {}^0{\dot{\bm{\omega}}}_j+{}^0{\bm{\omega}}_j\times({\bm{I}}_j {}^0{\bm{\omega}}_j)
 \end{aligned}
 \end{equation}
 
 The dynamic model is expressed in matrix form, where the thrust force of all quadrotors $\bm{f}={\begin{bmatrix} {\bm{f}}_1^T {\bm{f}}_2^T & \cdots & {\bm{f}}_n^T \end{bmatrix}}^T$ are derived as (\ref{eq:DM3}), with the mass matrix  ${\bm{M}}_q$, and a wrench matrix $\bm{U}={\begin{bmatrix} {\bm{U}}_1^T & \cdots & {\bm{U}}_n^T \end{bmatrix}}^T$ such that ${\bm{U}}_j=\begin{bmatrix}
 \bm{0}& {}^0{\bm{R}}_p {}^p{\bm{u}}_i & \cdots &\bm{0}
 \end{bmatrix}$ has $s_j$ non-zero columns.
 \begin{equation}\label{eq:DM3}
 \bm{f} = {\bm{M}}_q {\ddot{\bm{q}}}_{\bm{a}} + \bm{U}\bm{t}+\bm{d}
 \end{equation}
 \begin{equation}\label{eq:d}\scriptsize
 \setlength{\arraycolsep}{1pt}
 {\bm{M}}_q=\begin{bmatrix}
 m_1 {\bf{I}}_3 \\
 & m_2{\bf{I}}_3 \\
 &&\ddots \\
 &&& m_n {\bf{I}}_3 \\
 \end{bmatrix} \text{ and }
  {\bm{d}}= \begin{bmatrix}
m_1\bm{g} \\ m_2\bm{g} \\ \vdots \\ m_n\bm{g}
\end{bmatrix}\normalsize
 \end{equation}
 
 The inverse dynamic model can be derived as (\ref{eq:IDM}) from (\ref{eq:DM1},\ref{eq:DM3},\ref{eq:2KM}).
\begin{equation}\label{eq:IDM}
{\bm{f}} = {\bm{D}}_q {\ddot{\bm{x}}}_{\bm{t}} + {\bm{G}}_q
\end{equation}
\begin{equation}
{\bm{D}}_q  = {\bm{M}}_q {\bm{B}}^{+} \bm{A} + \bm{U} {\bm{W}}^{+} \begin{bmatrix} {\bm{M}}_p & \bm{0} \end{bmatrix}
\end{equation}
\begin{equation}
{\bm{G}}_q = {\bm{M}}_q {\bm{B}}^{+} \bm{b} + \bm{U}{\bm{W}}^{+} \bm{c} + \bm{d}
\end{equation}
 
\begin{figure}[!tpb]\centering
 \includegraphics[width=0.995\linewidth]{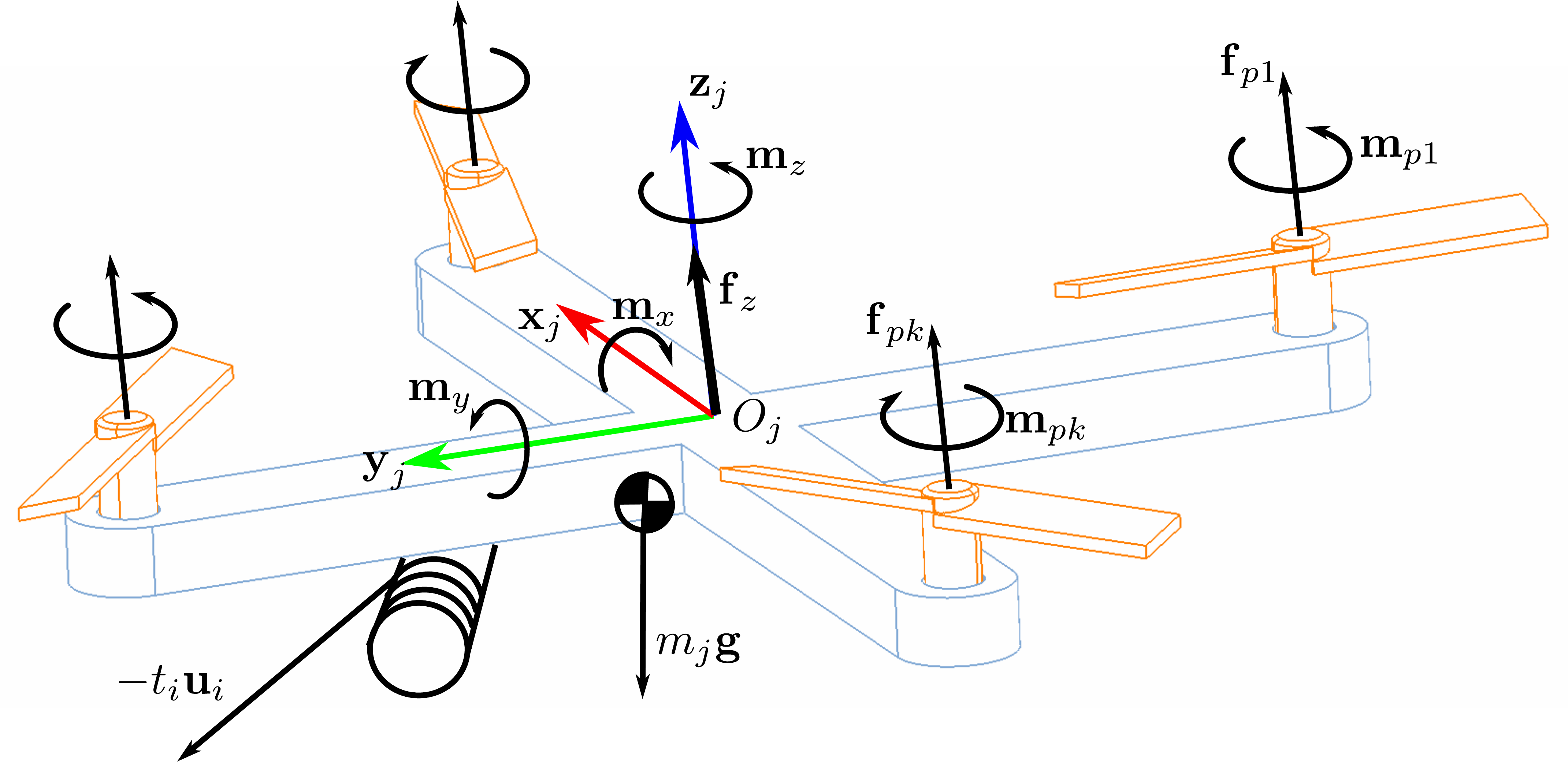}
 \caption{Free Body Diagram of the quadrotor $j$} \label{fig:quadrotor}
 \end{figure}

\added{Several types of quadrotors exist in the literature. In this paper, standard quadrotors with 4 co-planar propellers are considered.} For each quadrotor, the thrust force and thrust moment can be derived in the following way \cite{4SP}. As shown in Fig.~\ref{fig:quadrotor}, each propeller produces a thrust force $f_{p_k}$ and a drag moment $m_{p_k}$, for $k=1,2,3,4$. Considering that the aerodynamic coefficients are almost constant for small propellers, the following expression can be obtained after simplification:
 \begin{equation}
 f_{p_k}=k_f{\Omega_k}^2
 \end{equation}
 \begin{equation}
 m_{p_k}=k_m{\Omega_k}^2
 \end{equation}
 where $k_f$, $k_m$ are constants and $\Omega_k$ is the rotational rate of the $k^{\text{th}}$ propeller. \added{In our experiments, $k_f=\expnumber{3.55}{-6}$~$N\cdot s^2/{Rad}^2$ and $k_m=\expnumber{5.4}{-8}$~$N\cdot s^2/{Rad}^2$ were empirically identified using an free-flight identification method proposed by some of the authors, currently under review.}
 
 The total thrust force and moments expressed in the $j^{\text{th}}$ quadrotor frame $\mathcal{F}_j$ are $\setlength{\arraycolsep}{1.5pt}{\begin{bmatrix}0&0&f_z\end{bmatrix}}^T$ and $\setlength{\arraycolsep}{1.5pt}{\begin{bmatrix}
 m_x&m_y&m_z
 \end{bmatrix}}^T$, respectively, where $r$ is the distance between the propeller and the center of the  quadrotor. Importantly, we could indicate that quadrotors have four DOFs corresponding to roll, pitch, yaw and upward motions.
 \begin{equation}\label{eq:quad}
 \begin{bmatrix}
 f_z\\m_x\\m_y&\\m_z
 \end{bmatrix}=\begin{bmatrix}
 1&1&1&1\\
 0&r&0&-r \\ -r&0&r&0 \\ -\frac{k_m}{k_f}&\frac{k_m}{k_f}&-\frac{k_m}{k_f}&\frac{k_m}{k_f}
 \end{bmatrix}\begin{bmatrix}
 f_{p_1} \\ f_{p_2} \\ f_{p_3} \\ f_{p_4}
 \end{bmatrix}
 \end{equation}

\section{Centralized controller design}\label{sec:control}

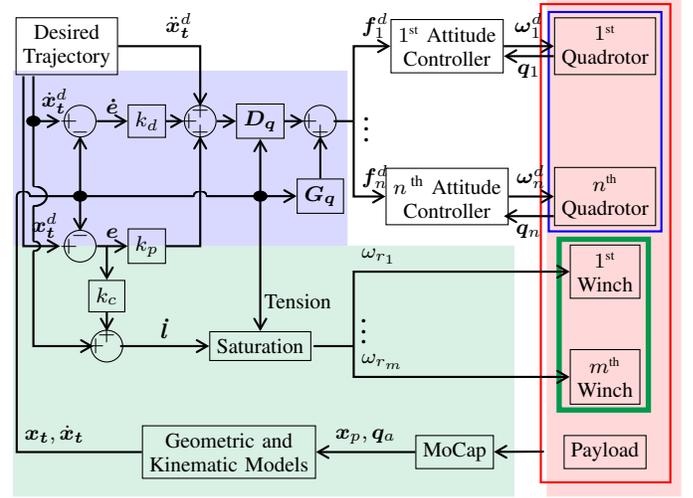
\begin{figure}[!tpb]\centering
 \tikzstyle{process} = [rectangle, minimum width=0.2cm, minimum height=0.2cm, text centered, draw=black]
 \tikzstyle{arrow} = [->,>=angle 60,thick]
 \begin{tikzpicture}[node distance=2cm]\footnotesize
 \node[anchor=south west,inner sep=0] (image) at(0,0){\includegraphics[width=1\linewidth]{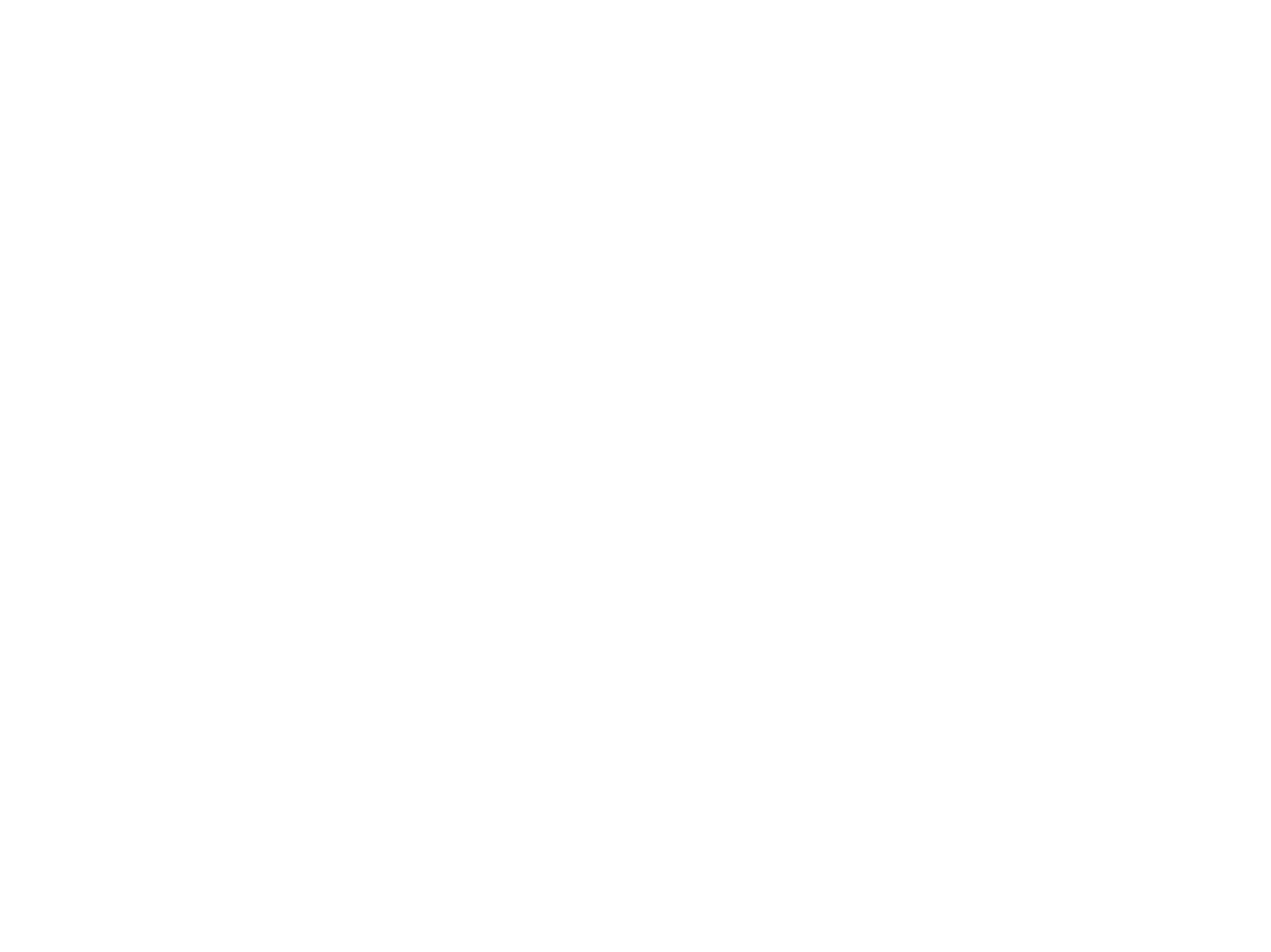}};
  \begin{scope}[x={(image.south east)},y={(image.north west)}]
   \definecolor{mygreen}{rgb}{0,0.6,0.3};
   \draw[fill=blue!15,blue!15] (0.0,0.45) rectangle (0.5,0.85);
   \draw[fill=mygreen!15,mygreen!15] (0.0,0.0) rectangle (0.75,0.5);
   \draw[fill=red!15,red!15] (0.8,0.0) rectangle (1.0,1.0);
   \node[process,align = center](DT) at (0.08,0.9) {Desired \\ Trajectory};
   \node[process,align = center](AC) at (0.65,0.9) {$1^{\text{\tiny st}}$ Attitude \\ Controller};
   \node[process, right of = AC, xshift=0.1cm, align = center](qr){$1^{\text{\tiny st}}$ \\ Quadrotor};
   \node[process, below of = AC, align = center](ACn) {$n^{\text{ th}}$ Attitude \\ Controller};
   \node[process, right of = ACn, xshift=0.1cm, align = center](qrn){$n^{\text{\tiny th}}$ \\ Quadrotor};
   \node[process, below of = qrn, yshift=1cm, align = center](w1) {$1^{\text{\tiny st}}$ \\ Winch};
   \node[process, below of = w1,yshift=0.6cm, align = center](wm){$m^{\text{\tiny th}}$ \\ Winch};
   \node[process, below of = wm,yshift=1cm, align = center](pay){Payload};
   \node[process, left of = pay,align = center](mocap){MoCap};
   \node[process, left of = mocap, xshift=-1cm, align = center](GKM){Geometric and \\ Kinematic Models};
   \node[process,align = center](D) at (0.2,0.75) {$k_d$};
   \node[process,align = center](P) at (0.2,0.5) {$k_p$};
   \node[process,align = center](C) at (0.14,0.4) {$k_c$};
   \node[process, align = center](DFC) at (0.37,0.75) {$\bm{D_q}$};
   \node[process, align = center](TDA) at (0.46,0.6) {$\bm{G_q}$};
   \node[process, align = center](norm) at (0.37,0.3) {Saturation};
   \tikzset{
    charge node/.style={inner sep=0pt},
    pics/sum block/.style n args={4}{
      code={
        \path node (n) [draw, circle, inner sep=0pt, minimum size=12pt] {}
          (n.north) +(0,-2pt) node [charge node] {$#1$}
          (n.south) +(0,2pt) node [charge node] {$#2$}
          (n.west) +(2pt,0) node [charge node] {$#3$}
          (n.east) +(-2pt,0) node [charge node] {$#4$}
          ;
      }
    }
  }
   \path pic at (0.1,0.75) {sum block={}{-}{+}{}};
   \path pic at (0.28,0.75) {sum block={+}{+}{+}{}};
   \path pic at (0.46,0.75) {sum block={}{+}{+}{}};
   \path pic at (0.1,0.5) {sum block={-}{}{+}{}};
   \path pic at (0.14,0.3) {sum block={+}{}{+}{}};
   \draw[red,thick] ($(qr.north west)+(-5pt,5pt)$)  rectangle ($(pay.south east)+(9pt,-5pt)$);
   \draw[blue,thick] ($(qr.north west)+(-2pt,2pt)$)  rectangle ($(qrn.south east)+(2pt,-2pt)$);
   \draw[mygreen,line width=2pt] ($(w1.north west)+(-4pt,2pt)$)  rectangle ($(wm.south east)+(3pt,-2pt)$);
   \draw[arrow](DT)-|node [above left] {${{\ddot{\bm{x}}}_{\bm{t}}}^d$}(0.28,0.78);
   \draw[arrow](0.03,0.84)|-node [above right] {${{\dot{\bm{x}}}_{\bm{t}}}^d$}(0.07,0.75);
   \draw[arrow](0.015,0.84)|-node [above right] {${\bm{x}}_{\bm{t}}^d$}(0.07,0.5);
   \draw[arrow](0.125,0.75)--node [above] {$\bm{\dot{e}}$}(D);
   \draw[arrow](D)--(0.255,0.75);
   \draw[arrow](0.305,0.75)--(DFC);
   \draw[arrow](DFC)--(0.435,0.75);
   
   \draw[arrow](0.795,0.089)--(mocap);
   \draw[arrow](mocap)--node[above]{${\bm{x}}_p, {\bm{q}}_a$}(GKM);
   \draw[thick] (GKM)-|node[above right]{$\bm{x_t},{\dot{\bm{x}}}_{\bm{t}}$}(0.005,0.6);
   \draw[arrow](0.025,0.6)-|(0.1,0.72);
   \draw[arrow](0.1,0.6)--(0.1,0.53);
   \begin{scope}
    \clip (0.005,0.6) rectangle (0.025,0.61);
    \draw[thick] (0.015,0.6) circle(0.01);
   \end{scope}
   \draw[arrow](0.125,0.5)--node[above]{$\bm{e}$}(P);
   \draw[arrow](P)-|(0.28,0.72);
   
   \draw[arrow](0.14,0.5)--(C);
   \draw[arrow](C)--(0.14,0.33);
   \draw[arrow](0.16,0.3)--node [above]{$\bm{\dot{l}}$}(norm);
   \draw[black,fill=black] (0.1,0.6) circle(0.01);
   \draw[thick](0.1,0.6)--(0.27,0.6);
   \begin{scope}
    \clip (0.27,0.6) rectangle (0.29,0.61);
    \draw[thick] (0.28,0.6) circle(0.01);
   \end{scope}
   \draw[arrow](0.29,0.6)-|(DFC);
   \draw[arrow](TDA)--(0.46,0.72);
   \draw[black,fill=black] (0.37,0.6) circle(0.01);
   \draw[arrow](0.37,0.6)--(TDA);
   \draw[arrow](0.37,0.6)--node[xshift=0.5cm,yshift=-0.5cm]{Tension}(norm);
   
   \draw[black,fill=black](0.03,0.75)circle(0.01);
   \draw[thick](0.03,0.75)--(0.03,0.62);
   \begin{scope}
    \clip (0.03,0.58) rectangle (0.32,0.62);
    \draw[thick] (0.03,0.6) circle(0.02);
   \end{scope}
   \draw[thick](0.03,0.58)--(0.03,0.51);
   \begin{scope}
    \clip (0.03,0.49) rectangle (0.31,0.51);
    \draw[thick] (0.03,0.5) circle(0.01);
   \end{scope}
   \draw[arrow](0.03,0.49)|-(0.12,0.3);
   
   \draw[thick](0.48,0.75)--node[xshift=0.3cm]{$\bm{\vdots}$}(0.51,0.75);
   \draw[arrow](0.51,0.75)|-node[above right]{${\bm{f}}_1^d$} (AC);
   \draw[arrow](0.51,0.75)|-node[above right]{${\bm{f}}_n^d$}(ACn);
   \draw[thick](0.45,0.3)--node[xshift=0.4cm,yshift=0.3cm]{$\bm{\vdots}$}(0.51,0.3);
   \draw[arrow](0.51,0.3)|-node[above right]{${\omega}_{r_1}$} (w1);
   \draw[arrow](0.51,0.3)|-node[above right]{${\omega}_{r_m}$}(wm);
   \draw[arrow](AC)--node [above] {${\bm{\omega}}_1^d$}(qr);
   \draw[arrow](0.81,0.88)--node [below]{${\bm{q}}_1$}(0.735,0.88);
   \draw[arrow](ACn)--node [above] {${\bm{\omega}}_n^d$}(qrn);
   \draw[arrow](0.81,0.56)--node [below]{${\bm{q}}_n$}(0.735,0.56);
 \end{scope}  
 \end{tikzpicture}
 \definecolor{mygreen}{rgb}{0,0.6,0.3}
 \caption{\added{The control diagram showing the desired thrust of quadrotors ({\color{blue}{blue}}), the control loop of winches ({\color{mygreen}{green}}) and the VACTS plant ({\color{red}{red}})}}
 \label{flowchart:vactscontroller}
 \end{figure}

The overview of a centralized feedback linearization VACTS control diagram is shown in Fig.~\ref{flowchart:vactscontroller}, which includes the control loop of quadrotors, as well as winches. Once the thrusts and the cable velocities are decided by the PD control law (\ref{eq:vactscontrollaw1}) and the P control law (\ref{eq:vactscontrollaw2}), respectively, the quadrotors can be stabilized by attitude controller and the cable lengths can be actuated by servo motors. The poses of quadrotors and the payload can be tracked by the Motion Capture System (MoCap). Therefrom the task space state vector $\bm{x_t}$ and its derivative are derived based on the geometric and kinematic models.
\subsection{Desired Thrust}
 \added{The desired thrust represents the reference thrust generation of quadrotors for sharing burden.} The control law is proposed based on the inverse dynamic model of the VACTS. \added{This control law for a regular parallel robot guarantees asymptotic convergence is the absence of modelling errors~\cite{paccot2009review}. As the error behaves as a second-order system, the gains can be determined by a cutoff frequency and a damping ratio.}
\begin{equation}\label{eq:vactscontrollaw1}
{\bm{f}}^d = {\bm{D}}_q({{\ddot{\bm{x}}}_{\bm{t}}}^d + k_d({{\dot{\bm{x}}}_{\bm{t}}}^d-{\dot{\bm{x}}}_{\bm{t}})  + k_p({\bm{x}}_{\bm{t}}^d-{\bm{x_t}}) ) +{\bm{G}}_q
\end{equation}
\subsection{Velocity Control Loop}
The velocity control loop represents the control loop of winches for actuating cable lengths. ${\dot{\bm{l}}}^d$ is obtained from the desired trajectory. Moreover, the function of saturation in the control diagram will be introduced in detail. Let us assume that there is a linear relationship between the maximum speed of winch ${\omega}_{r_{max}}$ and its torque, which is called the speed-torque characteristics. For a certain torque, the servo motor can tune the winch speed from $-{\omega}_{r_{max}}$ to $+{\omega}_{r_{max}}$. Therefore, the output signal can be determined based on the knowledge of cable tension and desired cable velocity. Furthermore, the output signal is limited within the $70\%$ of its working range for ensuring safety.
\begin{equation}\label{eq:vactscontrollaw2}
 \dot{\bm{l}}={\dot{\bm{l}}}^d+k_c({\bm{l}}^d-\bm{l})
 \end{equation}
\subsection{Attitude Control}
The aim of the attitude controller \cite{ETH} is to provide the desired thrust force and stabilize the orientation of quadrotor. \added{An estimated attitude ${\bm{q}}_j$ in quaternion representation is used as the feedback signal.} The yaw angle is set to be a given value such as $0\degree$ in order to fully constrain the problem\added{, however it may be interesting in the future to optimize it as a function of the state to reduce moments applied by the cable to the quadrotor}. In \cite{lee2010geometric}, a second-order attitude controller with feed-forward moment is proposed, \added{to which a feed forward moment from the desired cable tension could be applied}.
 
\section{Performance evaluation}\label{sec:wrench}

\subsection{Wrench Analysis}

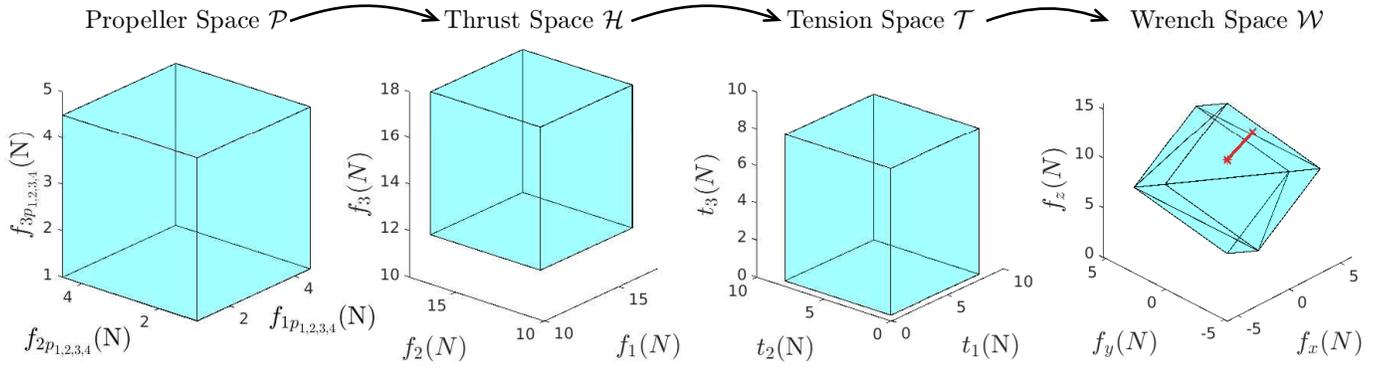
\begin{figure*}[!tpb]\centering
	\begin{tikzpicture}
	\node[anchor=south west,inner sep=0] (image) at(0,0){\includegraphics[trim={3.5cm 0 3.5cm 0},clip, width=1\linewidth]{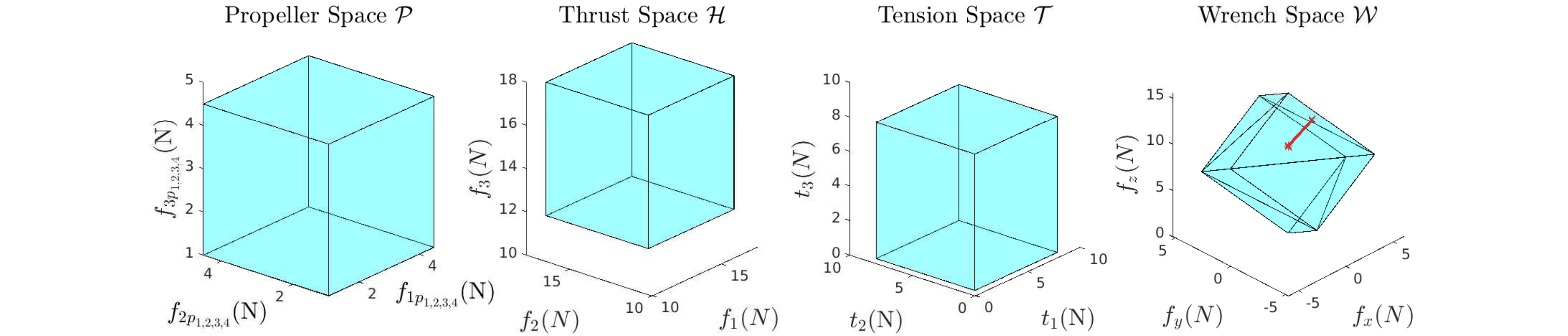}};
	\begin{scope}[x={(image.south east)},y={(image.north west)}]
	\draw[->,>=angle 45,line width=1pt](.21,.95) to [out=20,in=160](.32,.95);
	\draw[->,>=angle 45,line width=1pt](.46,.95) to [out=20,in=160](.57,.95);
	\draw[->,>=angle 45,line width=1pt](.72,.95) to [out=20,in=160](.81,.95);
	\end{scope}
	\end{tikzpicture}
	\caption{\added{The four spaces of wrench analysis for a VACTS}}
	\label{fig:4space}
\end{figure*}

The available wrench set ${\mathcal{W}}_a$ is the set of wrenches that the mechanism can generate and depends on the state. ${\mathcal{W}}_a$ can be obtained by three space mappings, which is an extension of that introduced in \cite{wrench} for a general classical ACTS. This paper details the differences, which is caused 1) by the winch preventing cables from passing through the COM of the quadrotor\added{, and 2) by the winch torque limits}.

\subsubsection{Propeller space}

Generally, for an UAV with $n_p$ propellers, its propeller space $\mathcal{P}$ can be determined by the lower \added{(to prevent the motors from stalling)} and upper bounds of thrust force that each propeller can generate. Considering $n$ UAVs, the propeller space is (\ref{eq:propellerspace}).
 \begin{equation}\label{eq:propellerspace}
 \mathcal{P}=\left\{{\bm{f}}_p\in{\rm I\!R}^{(n\cdot n_p)}: \underline{{\bm{f}}_p}\leq{\bm{f}}_p\leq\bar{{\bm{f}}_p}\right\}
 \end{equation}
 
 For a certain set of cable tensions, the compensation moment can be obtained according to (\ref{eq:quadmoment}), which is risen from non-coincident COM and geometric center of quadrotors, as well as the cables not passing through the geometric center. Then the maximum value of thrust force of the $j^{\text{th}}$ quadrotor can be found by analyzing an optimization problem as (\ref{eq:optimf}).
 \begin{equation}\label{eq:optimf}
 \bar{f_z}=\underset{f_{p_k}\in \mathcal{P}}{\arg\max} f_z \text{ s.t. (\ref{eq:quad})}
 \end{equation}
  \added{This method accounts for individual propeller saturation and may be applied to any (V)ACTS, for any co-planar multi-rotor type UAV.}

 \subsubsection{Thrust, tension and wrench spaces}
 
 The thrust space $\mathcal{H}$ is obtained by space mapping from the propeller space $\mathcal{P}$ as~(\ref{eq:optimf}). It represents the set of available thrusts of quadrotors. The tension space $\mathcal{T}$ shows the available cable tensions, which considers not only the space mapping from the thrust space, but also the torque capacity of winch. The available wrench set ${\mathcal{W}}_a$ can be obtained by mapping tension space $\mathcal{T}$ to wrench space $\mathcal{W}$~\cite{wrench} by using the convex hull method~\cite{wrenchbasic}. 
 
 The capacity margin $\gamma$ is a robustness index that is used to analyse the degree of feasibility of a configuration \cite{arachnis}. It is defined as the shortest signed distance from \added{the task wrench} ${\mathcal{W}}_t$ \added{(here $\mathcal{W}_t = {m_p{\bf g}}$)} to the boundary of the ${\mathcal{W}}_a$ zonotope. The payload mass, the cable directions and the capability of UAVs and winches all affect the value of capacity margin.
 
 For example, these four spaces can be presented visually as Fig.~\ref{fig:4space} for the VACTS with three quadrotors, three winches and cables, and a point-mass \added{however the method is general for all ACTS designs \cite{wrench}. Note that $\mathcal{P}\in\mathbb{R}^{12}$ is shown as three dimensional, although in reality it is a 12 dimensional hypercube (as all motors and propellers are identical).}
 
 \subsection{Comparison}
 
  \begin{table}[!tpb]
\caption{wrench analysis: general parameters}
\label{table:generalparam}
\begin{center}
\vspace{-1em}\begin{tabular}{|c||c|c|}
  \hline
  symbol & physical meaning & value \\
  \hline
  $m_q$ & the mass of quadrotor only & 1.05kg \\
  $m_w$ & the mass of winch only & 150g \\
  $m_p$ & the mass of payload & 1kg \\
  $\bar{f_p}$ & the maximum thrust of propellers & 4.5N\\
  ${}^j{\bm{x}}_Q$ & the COM of quadrotor in $\mathcal{F}_j$ & ${\begin{bmatrix}0 & 0 & -2\end{bmatrix}}^T$cm \\
  ${}^w{\bm{x}}_A$ & the COM of winch in $\mathcal{F}_{wi}$ & ${\begin{bmatrix}0 & 0 & 0\end{bmatrix}}^T$cm \\
  ${\bar{\tau}}_{w}$ & the maximum torque of winches & 6kg$\cdot$cm \\
  $\bm{\phi}$ & the azimuth angles of quadrotors & $\begin{bmatrix} 0 & 120 & -120\end{bmatrix}\degree$ \\
  \hline
\end{tabular}
\end{center}
\end{table}

The most notable difference of wrench analysis between ACTS and VACTS is that $\mathcal{H}$ of the ACTS is constant, while $\mathcal{P}$ of the VACTS is constant and $\mathcal{H}$ is variable.
 
\begin{figure*}[!thpb]\centering
  \includegraphics[width=.9\linewidth]{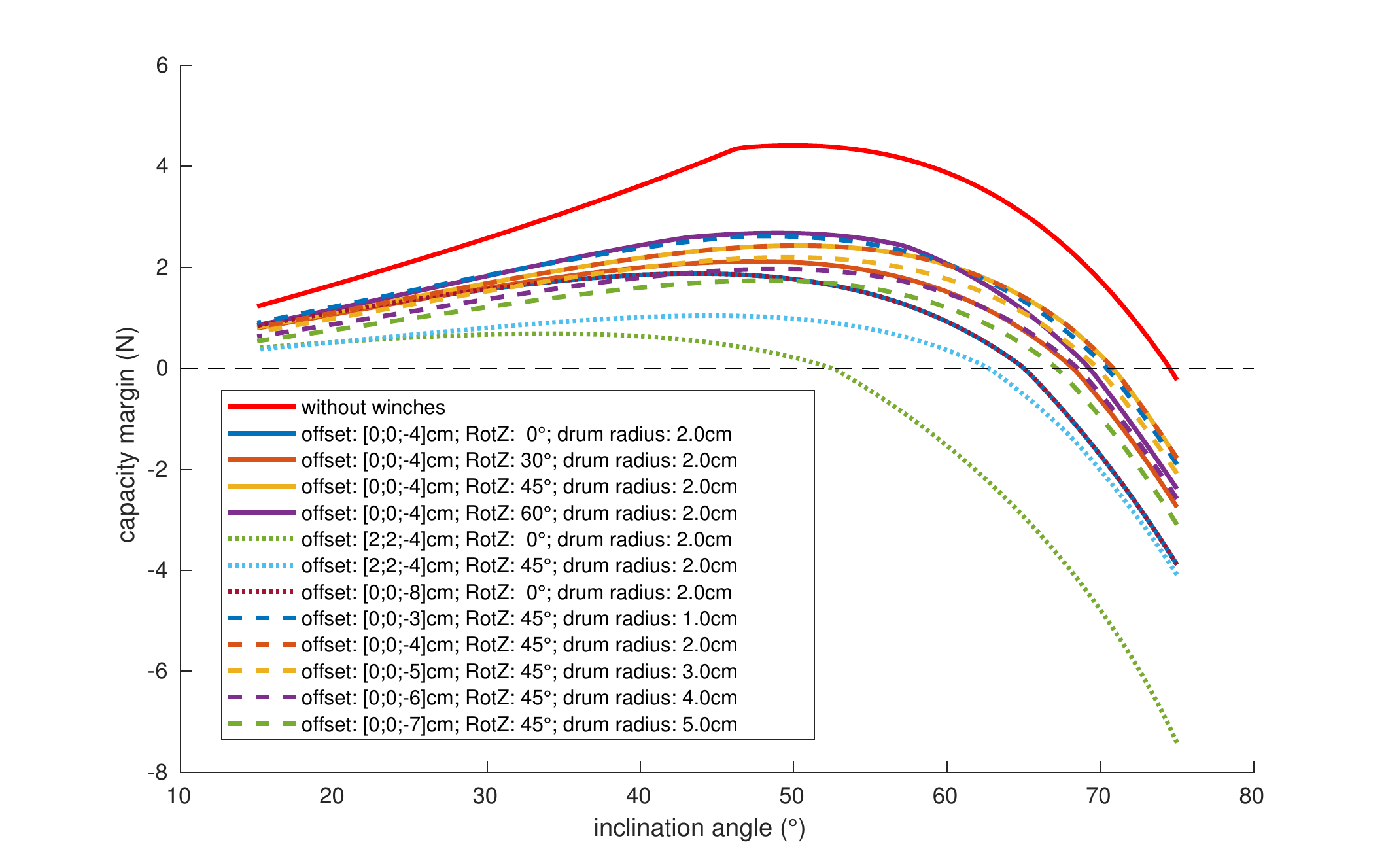}
 \vspace{-1em} \caption{Capacity margin for different mounted ways and drum radii with the offset along drum $2cm$ of the coincident point of the $i^{\text{th}}$ cable-winch pair in the winch frame ${}^w{\bm{x}}_{Ii}$. The "offset" in the legend represents the translation vector ${}^j{\bm{x}}_w$ between the winch frame and the quadrotor frame. The "$RotZ$" represents the rotation angle around $z$ axis between the winch frame and the quadrotor frame.}
 \label{fig:case1}
 \end{figure*}

The case study parameters  for a VACTS with three quadrotors, three cables and winches, and one point-mass are shown in Table~\ref{table:generalparam}. \added{Figure~\ref{fig:case1}} shows the effect of different parameters related to winch (drum radius, offset, and orientation). From this case study, we conclude that (i) For the same system configuration, the VACTS needs extra energy to compensate the moment generated by cable tension because of the offset of winch. Therefore the winch should be mounted to the quadrotor as close to its centroid as possible; (ii) Additionally, the drum radius should be set less than a certain value according to the simulation results, such as $2cm$ for this case study; (iii) The optimal inclination angle is different after embedding winches; (iv) Last but not least, the servo motor selection is also under constraints. The stall torque of servo motor should be large enough to support the force transmission.
 
 The performance of ACTS and VACTS can be evaluated from different points of view: ($i$)~For the same system configuration for ACTS and VACTS, the VACTS needs extra energy to compensate the moment generated by cable tensions. Therefore the VACTS has a smaller thrust space than the ACTS. So we can come up with the conclusion that the ACTS has a better performance in terms of available wrench set than the VACTS in unconstrained environments; 
 
 ($ii$)~In constrained environments, maybe the ACTS can not achieve the optimal system configuration because of its large size. For example, the ACTS can not achieve the optimal  inclination angle ($50\degree$) because of the big width of the overall system which can not be implemented in an environment with limited width.  While the VACTS can reshape its size and achieve an optimal configuration. Therefore, the VACTS can behave better than the ACTS in constrained environments and it will reshape the available wrench set; 
 
\begin{figure}[!pb]\centering
   \includegraphics[width=.69\linewidth]{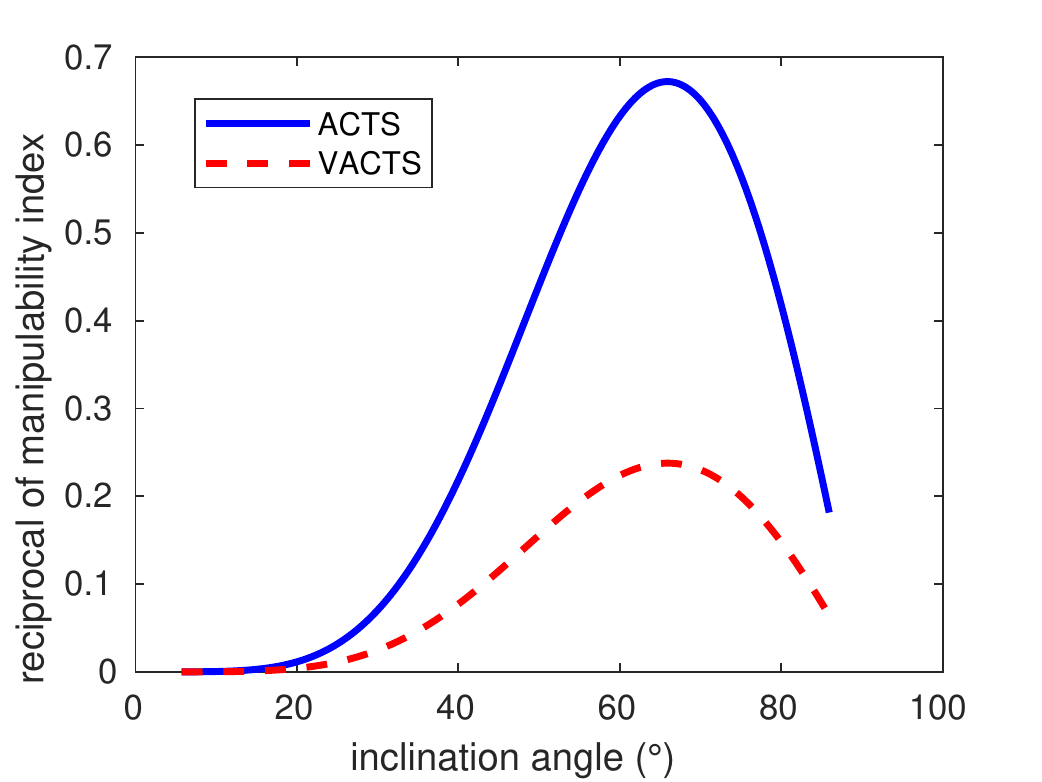}
\vspace{-1em} \caption{The reciprocal of the manipulability index for ACTS and VACTS with three quadrotors and a point-mass. The lower, the better.}
 \label{fig:invw}
   \end{figure}

 ($iii$)~The VACTS has a better manipulability than ACTS from the fact that the VACTS has a larger velocity capacity along cable directions than ACTS because of the actuated cable lengths, which can also be proven by simulation results as Fig.~\ref{fig:invw}. The manipulability index \cite{stock2003optimal} $w_s$ as defined in~(\ref{eq:manipulability}) is proportional to the volume of the ellipsoid of instantaneous velocities and that it can be calculated as the product of the singular values of the normalized Jacobian matrix ${\bm{J}}_{norm}$. For example, for the VACTS with three quadrotors, three cables and winches, and one point-mass, ${\bm{J}}_{norm}$ is derived as~(\ref{eq:normJ}) considering the actuated cable lengths in joint space rather than task space;
\begin{equation}\label{eq:manipulability}
w_s = \sqrt{det({\bm{J}}_{norm}{\bm{J}}_{norm}^T)}\text{, or } w_s = {\lambda}_1{\lambda}_2\cdots{\lambda}_r
\end{equation}
\begin{equation}\label{eq:normJ}
{\bm{J}}_{norm} = diag(1,1,1,l_1,l_1,l_2,l_2,\cdots,l_m,l_m) \bm{J}
\end{equation}
 (iv) On one hand, the actuated cable lengths increase the control complexity from $4n$ to $4n$+$m$. On the other hand, they improve the reconfigurability from $3n$-$m$ to $4n$-$m$ at the same time, which means the system is more flexible. Specifically, each quadrotor has four control variables, while each actuated cable length has one control variable. Therefore the control complexity is $4n$ for an ACTS with $n$ quadrotors, while it’s $4n$+$m$ for a VACTS with $n$ quadrotors and $m$ cables. In the meanwhile, a single cable has two reconfigurable variables (azimuth and inclination angles) and a pair of coupled cables has one reconfigurable variable (inclination angle). In an ACTS with $n$ quadrotors and $m$ cables, there are $2n$-$m$ single cables and $m$-$n$ pairs of coupled cables. So an ACTS has $3n$-$m$ reconfigurable variables. If the cable lengths are actuated, extra $n$ independent reconfigurable variables appear, i.e., a VACTS has $4n$-$m$ reconfigurable variables.
 
\section{Experimental results}\label{sec:exp}
  
\begin{figure}[!thpb]\centering
 \begin{tikzpicture}
 \node[anchor=south west,inner sep=0] (image) at(0,0){\includegraphics[width=1\linewidth]{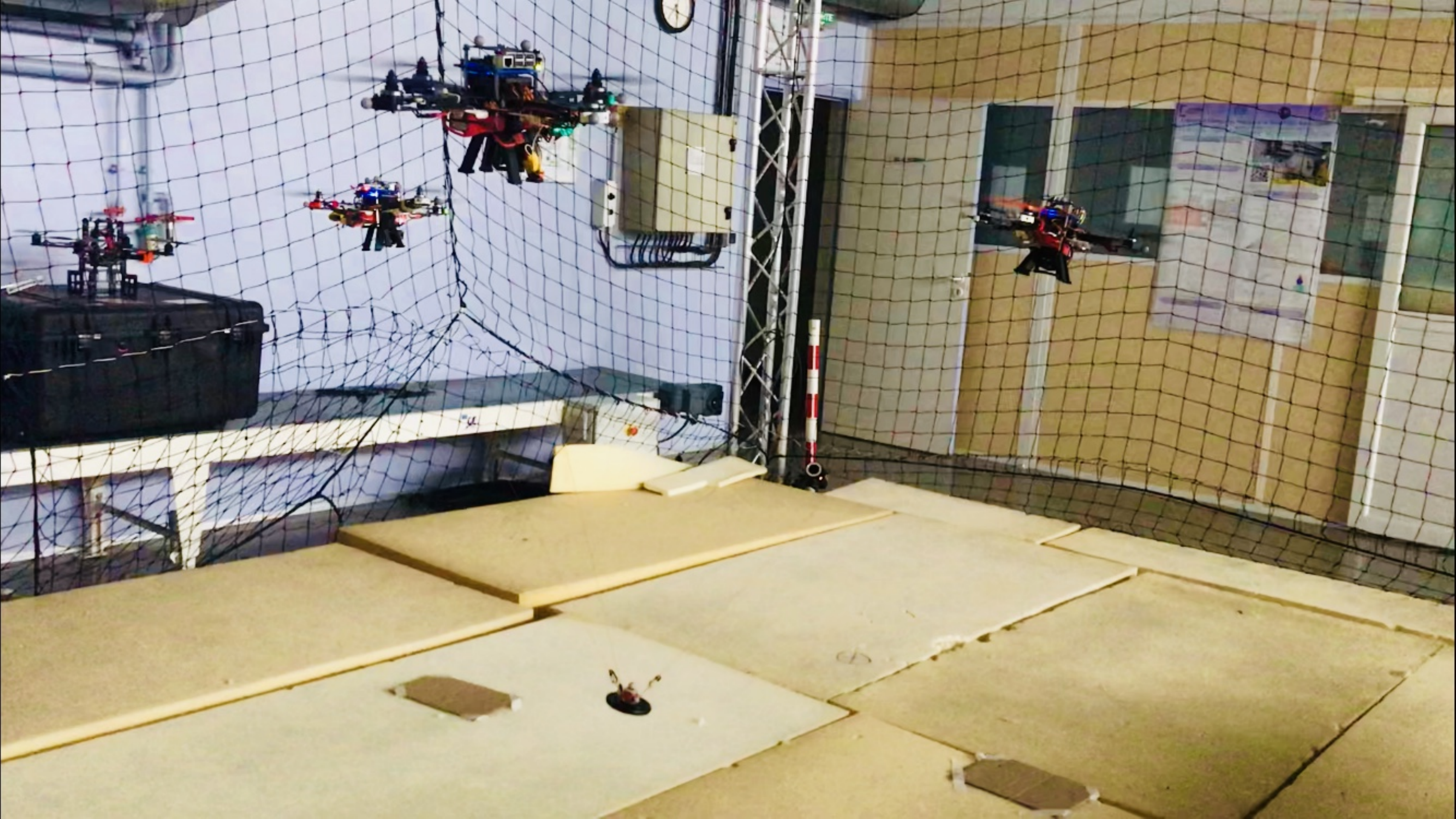}};
  \begin{scope}[x={(image.south east)},y={(image.north west)}]
   \definecolor{mygreen}{rgb}{0,0.5,0.3};
   \draw[mygreen,->,>=angle 60](0.5,0.1)--node [xshift=1.2cm] {point-mass}(0.45,0.12);
   \draw[blue] (.56,.4) circle (0.3cm) node [xshift=0.8cm,yshift=-0.4cm] {MoCap};
   \draw[red](0.345,0.79)--(0.425,0.15);
   \draw[red](0.26,0.7)--(0.43,0.16);
   \draw[red](0.72,0.67)--(0.44,0.15);
   \end{scope}
 \end{tikzpicture}
 \caption{The VACTS prototype with 3 quadrotors, winches and a point-mass}\label{fig:prototype}
 \end{figure}
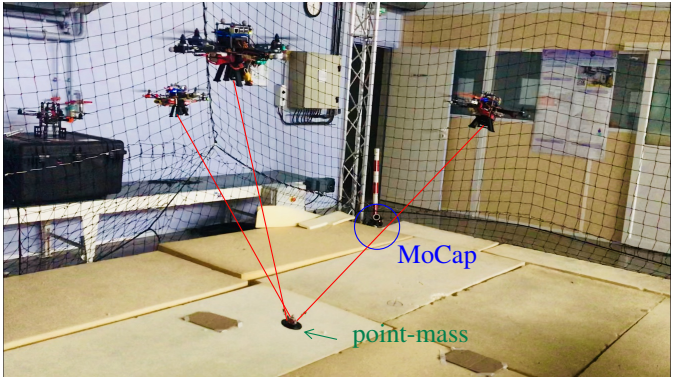

\begin{figure}[!thpb]\centering
  \begin{minipage}[t]{0.23\textwidth}\centering
  \includegraphics[width=1\linewidth]{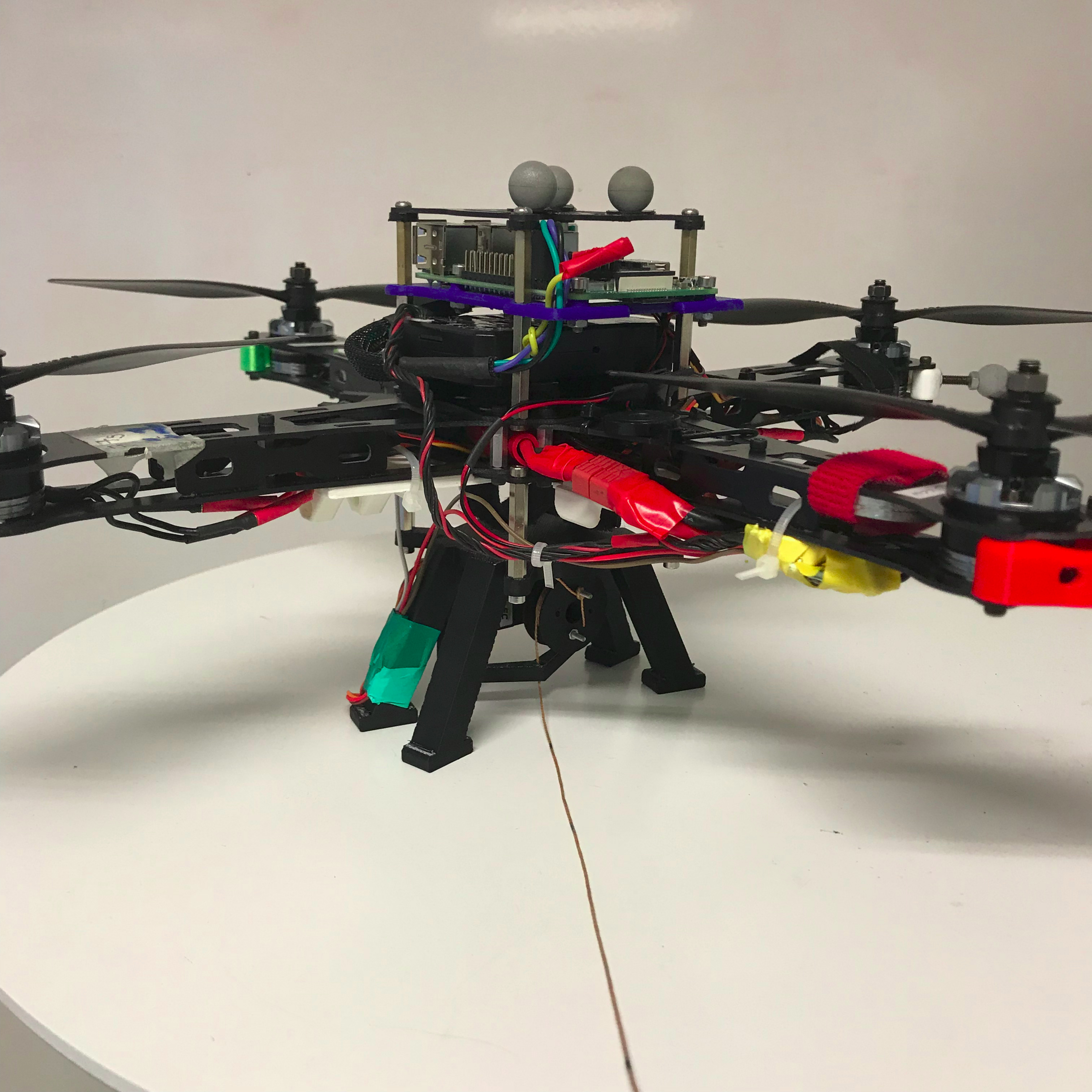}
  \caption{The quadrotor body with embedded winch for experiments} \label{fig:drone_exp}
  \end{minipage}
  \begin{minipage}[t]{0.23\textwidth}\centering
  \begin{tikzpicture}
 \node[anchor=south west,inner sep=0] (image) at(0,0){\includegraphics[width=1\linewidth]{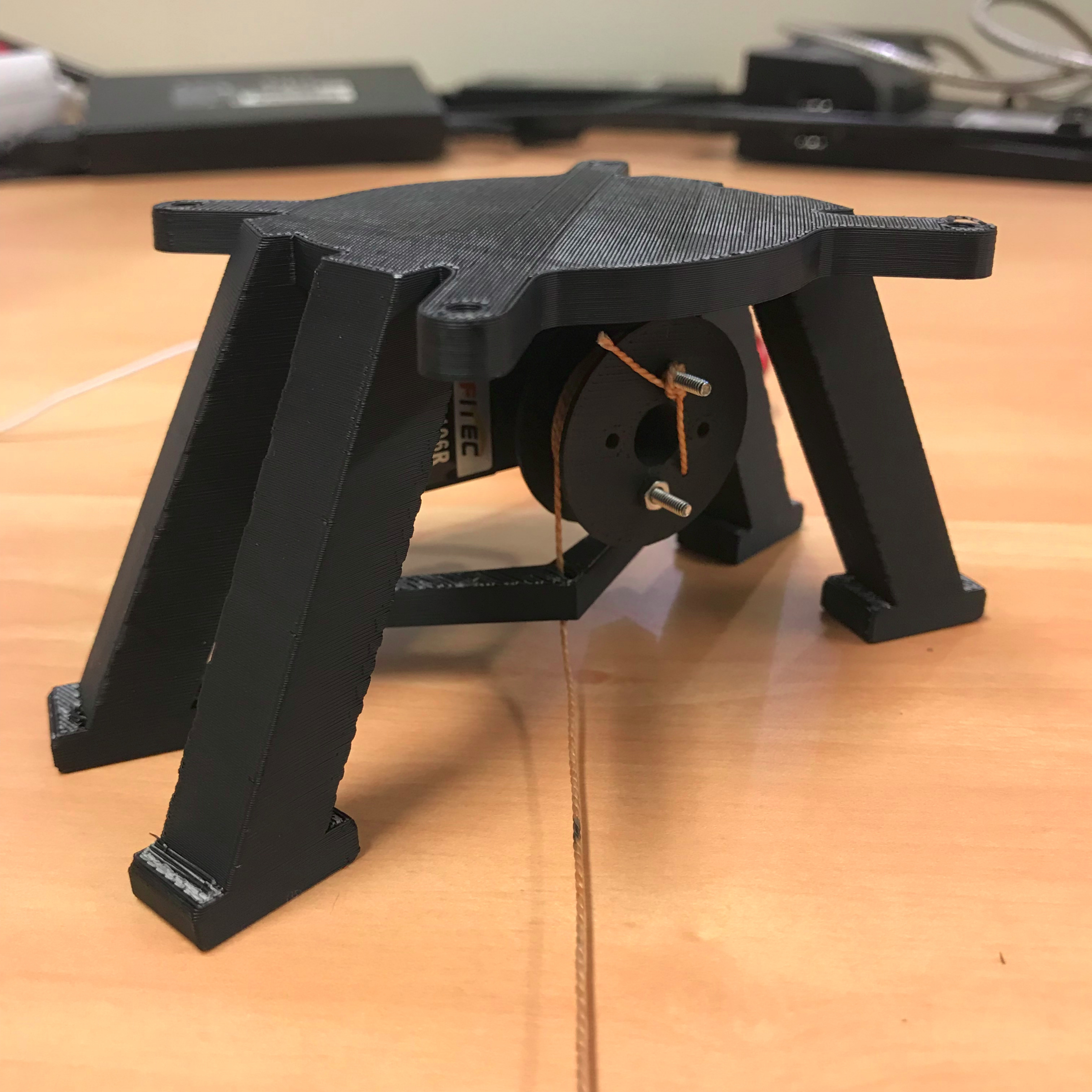}};
  \begin{scope}[x={(image.south east)},y={(image.north west)}]
   \draw[red,->,>=angle 60](0.6,0.3)--node[xshift=0.8cm,yshift=-0.7cm]{V-shape bar}(0.55,0.45);
   \end{scope}
 \end{tikzpicture}
  \caption{Winch prototype and support} \label{fig:winch_exp}
  \end{minipage}
  \end{figure}

The VACTS prototype with three quadrotors, winches and cables, and a point-mass $m_p=670g$ is shown in Fig.~\ref{fig:prototype}. The Quadrotor body is a Lynxmotion Crazy2fly as Fig.~\ref{fig:drone_exp} because of low cost and mass, while the embedded winch is shown in Fig.~\ref{fig:winch_exp}. The servo motor winch is FS5106R. The poses of quadrotors and the payload are tracked by the MoCap as we mentioned in Sec.~\ref{sec:control}. Therefrom the cable lengths and the unit vectors along cable directions are computed. \added{The hardware and software platform and their working frequencies are detailed in \cite{lastone}. The winch servo motor is controlled via ROS messages over Pixhawk AUX outputs.} The translation vector ${}^j{\bm{x}}_{Ii}$ is between the coincident point $I_i$ and the origin of the quadrotor frame, as shown in Fig.~\ref{fig:actsparam}. It is assumed to be a constant value \added{${[0,0,-6.4]}^T$cm}, thanks to the existence of guide hole in the V-shape bar of winch mount piece as shown in Fig.~\ref{fig:winch_exp}.

  \vspace{-2em}

The motion planning for a VACTS can be divided into several parts based on the task space state vector: the motion of payload, the system configuration of cable directions, and the cable lengths: (i) The motion of payload can be designed under the knowledge of environment; (ii) The design methodology for system configuration\added{\cite{lastone,caro2019}} varies from quasi-static to dynamic case. In short, it is designed under capacity margin and cost optimization; (iii) The cable lengths are determined on the basis of limited volume space. In order to prove that this novel system is feasible \added{and verify the ability of changing overall size}, a motion planning is considered as following. This system takes off firstly and then hover for several seconds. Then the cable lengths are changed from initial values to 1.4m, followed by decreasing from 1.4m to 1.0m by using a fifth-order polynomial trajectory planning method \cite{gasparetto2010optimal}. During the period of cable displacements, the desired payload position and system configuration are kept as constants. In other words, the quadrotors move along the cable direction while the payload stays at the same position.
 \begin{figure}[!tpb]\centering
 \includegraphics[width=1\linewidth]{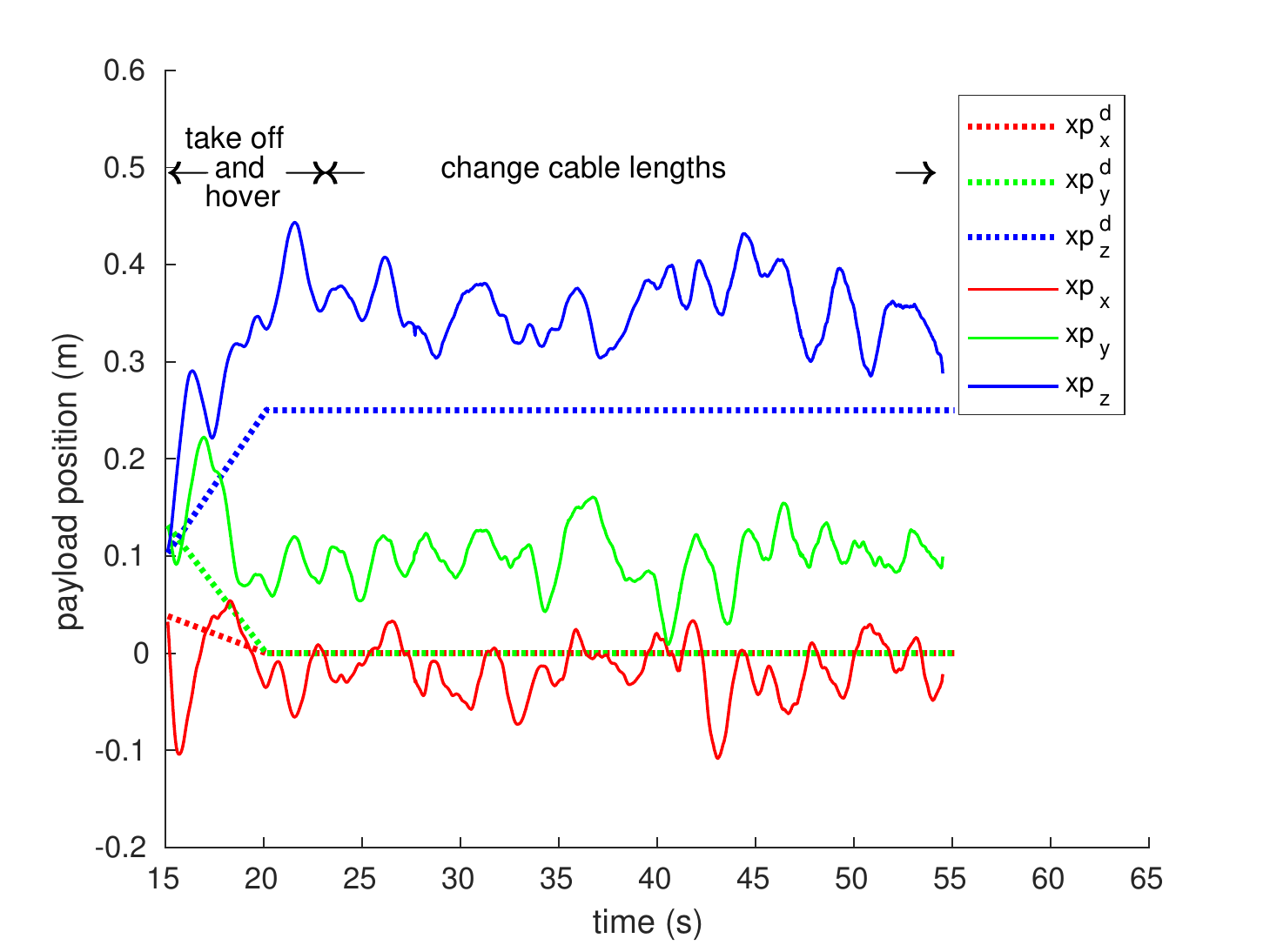}
 \caption{The payload tracking of VACTS with 3 quadrotors, 3 winches and a point-mass $m_p=670$~g}
 \label{fig:realtracking}
 \end{figure}
  \begin{figure}[!tpb]\centering
 \includegraphics[width=1\linewidth]{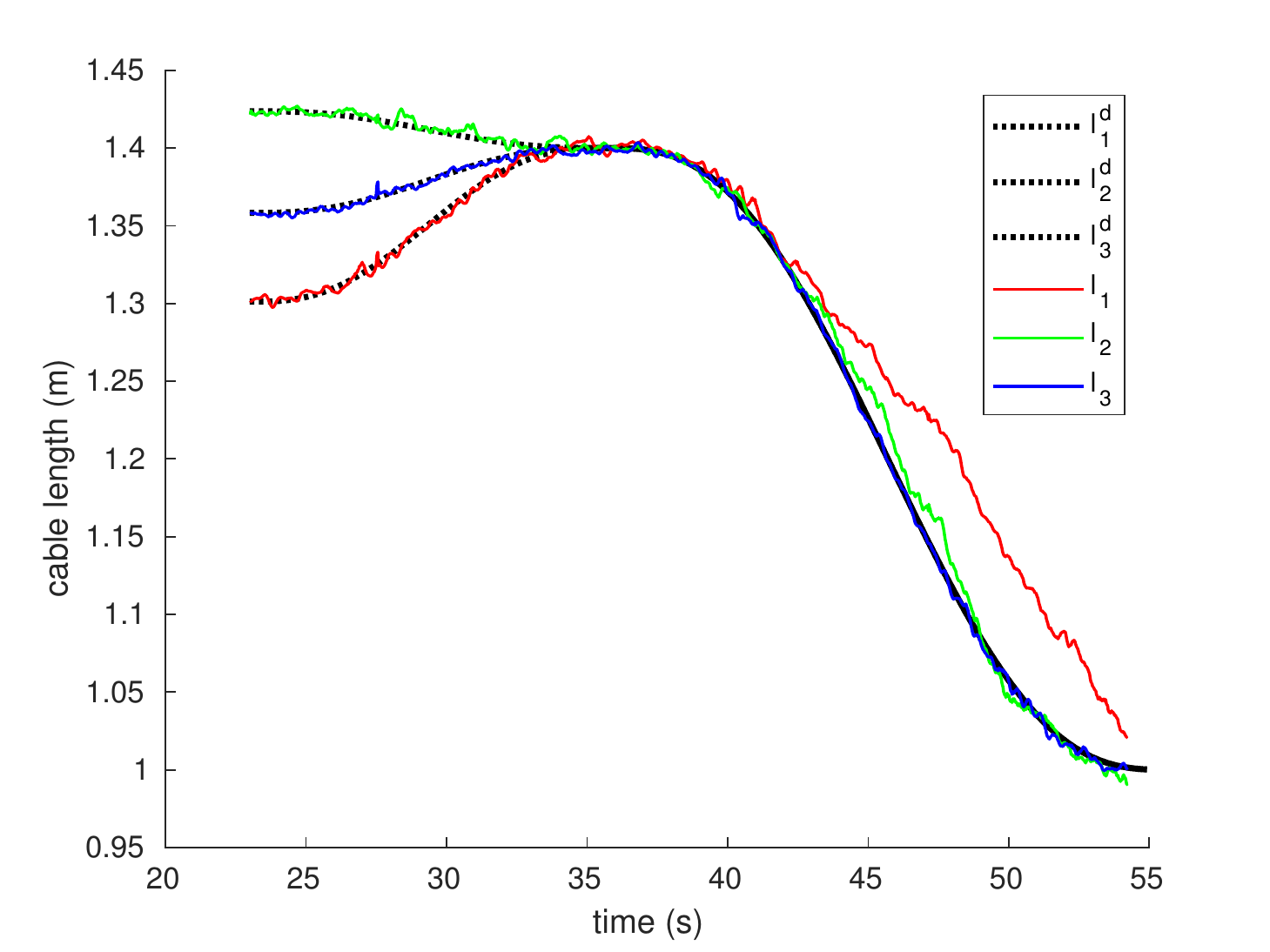}
 \vspace{-1em}\caption{The cable lengths tracking of VACTS with 3 quadrotors, 3 winches and a point-mass $m_p=670$~g}
 \label{fig:finalexp_l}
 \end{figure}
The experimental results of payload tracking and cable lengths tracking are shown in \added{Fig.~\ref{fig:realtracking} and Fig.~\ref{fig:finalexp_l}}. Moreover the corresponding video that shows the experimental results step by step can be found in the link\footnote{\href{https://drive.google.com/open?id=139fZmalOvZGxuETJf_h39mbEexZ2P5At}{\color{blue}{\nolinkurl{https://drive.google.com/open?id=139fZmalOvZGxuETJf_h39mbEexZ2P5At}}}}. From the experimental results, it's shown that the desired cable lengths are followed well and the system is stable under control. The payload tracking mean errors along $x$, $y$ and $z$ axis are $1.90cm$, $9.41cm$ and $10.67cm$, respectively. The standard deviations are $3.15cm$, $3.24cm$ and $3.29cm$, respectively. The cable lengths tracking mean errors for $l_1$, $l_2$ and $l_3$ are $-2.25cm$, $0.24cm$ and $0.10cm$, respectively. The standard deviations are $3.09cm$, $0.73cm$ and $0.23cm$, respectively. Even though there was a cable length that did not follow the trajectory well, it is acceptable. Specifically, the cable length velocity is limited by the winch capability. In the experiments, we define the limitation that the maximum speed of servo motor is 70\% of its actual maximum speed for security. At some point, the servo motor reaches its safety limit which leads to the small slope of $l_1$ ({\color{red}{red line}}) in Fig.~\ref{fig:finalexp_l}. \added{The error of payload tracking along $z$-axis is mainly due to  the low precision of attitude control for hovering or slow motion. The positioning error of the payload can be decreased by using the moment feedback in the attitude controller \cite{lee2010geometric}. The main control loop in Sec.~\ref{sec:control} runs at 50 Hz and the attitude controller runs at 200 Hz. The observed positioning errors are mainly due to this low frequency and the imperfect controller.}

Experimental results show that the VACTS is feasible and that the change of cable lengths can reshape the VACTS, which implies the possibility of passing through a constrained environment or limited space. Moreover, the precision of cable lengths control is much higher than the torque control of ACTS, which also implies the \added{(untested)} potential for improving precision of the VACTS when the payload position is fine-tuned by changing cable lengths.

\section{CONCLUSIONS}\label{sec:con}

This paper dealt with a novel aerial cable towed system with actuated cable lengths. Its non-linear models were derived and a centralized controller was developed. The wrench analysis was extended to account for propeller saturation resulting from the offset of the cable connections with the quadrotor and the winch was designed based on the wrench performance. The advantage of this novel system is in reshaping the overall size and wrench space in constrained environments, while it comes at the price of lower peak performance in unconstrained situations. The feasibility of the system is experimentally confirmed, showing the system to be capable of resizing while hovering.

Later on, ($i$)~a decentralized control method will be designed and applied in order to increase the flexibility and precision, with better fault tolerance; ($ii$)~cameras will be used to obtain the relative pose of the payload instead of motion capture system as in~\cite{bookquad} and \cite{vb} to develop a system that can be deployed in an external environment; ($iii$)~some criteria will be investigated to quantify the versatility of VACTS; ($iv$)~for the experimental demonstrations, a VACTS prototype with a moving-platform instead of point-mass will be developed to perform more complex aerial manipulations in a cluttered environment$^1$; ($v$)~a quadrotor attitude controller with feed forward moments from cable tension measurements or estimations will be implemented to improve performance.

\addtolength{\textheight}{-12cm}   

\bibliographystyle{IEEEtran}

\bibliography{root}

\end{document}